\PassOptionsToPackage{table}{xcolor}
\documentclass{article}
\pdfoutput=1
\usepackage{iclr2024_conference,times}

\usepackage[table]{xcolor}
\usepackage[utf8]{inputenc} %
\usepackage[T1]{fontenc}    %
\usepackage[hidelinks]{hyperref}         %
\usepackage{url}            %
\usepackage{booktabs}       %
\usepackage{amsfonts}       %
\usepackage{nicefrac}       %
\usepackage{microtype}      %
\usepackage{amsmath}
\usepackage{multirow}
\usepackage{subcaption}
\usepackage{graphicx}
\usepackage{cleveref}
\usepackage{algorithm}
\usepackage{algorithmic}
\usepackage{wrapfig}
\usepackage{color}
\usepackage{tcolorbox}
\usepackage{nicematrix}
\usepackage{enumitem}
\setlist{leftmargin=8mm}

\renewcommand{\paragraph}[1]{\noindent \textbf{#1}}
\usepackage{lipsum}

\newcommand\blfootnote[1]{%
  \begingroup
  \renewcommand\thefootnote{}\footnote{#1}%
  \addtocounter{footnote}{-1}%
  \endgroup
}

\title{Accurate Compression of Text-to-Image Diffusion Models via Vector Quantization}

\author{%
  Vage Egiazarian\textsuperscript{*}\footnotemark[1]\footnotemark[2]
  \And
  Denis Kuznedelev\textsuperscript{*}\footnotemark[1]\footnotemark[3]
  \And
  Anton Voronov\textsuperscript{*}\footnotemark[1]\footnotemark[2]\footnotemark[4]
  \And
  Ruslan Svirschevski\footnotemark[1]\footnotemark[2]
  \And
  Michael Goin\footnotemark[5]
  \And
  Daniil Pavlov\footnotemark[4]
  \And 
  Dan Alistarh\footnotemark[5]\footnotemark[6]
  \And
  Dmitry Baranchuk\footnotemark[1]
}

\iclrfinalcopy %
\begin{document}

\maketitle
{
\vspace{-7mm}
\footnotemark[1]Yandex Research \ \ \ \ \ \
\footnotemark[2]HSE University \ \ \ \ \ \
\footnotemark[3]Skoltech \ \ \ \ \ \
\footnotemark[4]MIPT \ \ \ \ \ \
\footnotemark[5]Neural Magic \ \ \ \ \ \
\footnotemark[6]IST Austria\\

\vspace{-3mm}
\hspace{35mm} \small{\url{https://yandex-research.github.io/vqdm}}
\vspace{3mm}
}

\begin{abstract}
  Text-to-image diffusion models have emerged as a powerful framework for high-quality image generation given textual prompts. 
  Their success has driven the rapid development of production-grade diffusion models that consistently increase in size and already contain billions of parameters.
  As a result, state-of-the-art text-to-image models are becoming less accessible in practice, especially in resource-limited environments.  
  \textit{Post-training quantization} (PTQ) tackles this issue by compressing the pretrained model weights into lower-bit representations.
  Recent diffusion quantization techniques primarily rely on \textit{uniform scalar quantization}, providing decent performance for the models compressed to $4$ bits.
  This work demonstrates that more versatile \textit{vector quantization} (VQ) may achieve higher compression rates for large-scale text-to-image diffusion models.
  Specifically, we tailor vector-based PTQ methods to recent billion-scale text-to-image models (SDXL and SDXL-Turbo), and show that the diffusion models of 2B+ parameters compressed to around $3$ bits using VQ exhibit the similar image quality and textual alignment as previous $4$-bit compression techniques.

\end{abstract}

\vspace{-12pt}
\section{Introduction}
\vspace{-7pt}

\blfootnote{$^*$ Equal contribution} In recent years, diffusion models~\citep{sohl2015deep,ho2020denoising,song2020score,rombach2022high} have revolutionized text-to-image (T2I) synthesis~\citep{saharia2022photorealistic, podell2024sdxl, pernias2023wuerstchen, rombach2021highresolution, esser2024scaling, betker2023improving}, transforming user textual prompts into highly realistic images. 
Motivated by trends in large language models (LLM), practitioners are eager to enhance the performance of text-to-image diffusion models by scaling and curating datasets~\citep{li2024scalability, kastryulin2024yaart, schuhmann2022laion5b} and increasing model sizes~\citep{esser2024scaling, podell2024sdxl, li2024scalability}. 
Current state-of-the-art models contain up to 12 billion parameters~\citep{flux}, with some studies suggesting that the performance gains from model scaling have yet to reach saturation~\citep{esser2024scaling}.
Therefore, we are likely to witness even larger open-source text-to-image models shortly.
However, the scaling trend of text-to-image models presages slow inference and large memory consumption in practice.
The problem gets more pronounced on edge devices with limited memory and weak processing units.
To address this issue, effective post-training diffusion model compression methods need to be developed.

Quantization~\citep{gholami2021survey} is a prominent compression technique that has gained significant attention due to its remarkable success in LLM compression~\citep{Frantar2023OPTQ, dettmers2023spqr, egiazarian2024extreme, tseng2024quip}.
 Weight quantization typically transforms the original parameter tensor into a condensed representation, where each weight or group of weights is replaced with a low-bit value. 
Effective quantization algorithms aim to minimize the discrepancy between the model outputs obtained with the original and compressed weights. 
Quantization methods can be broadly categorized into \emph{scalar quantization}, which projects each individual weight onto a one-dimensional grid, and \emph{vector quantization}~\citep{vq1, vq2}, which represents a group of weights as a vector from a \emph{codebook}.
Due to its simplicity, scalar quantization is widely adopted across different domains.
Several studies~\citep{li2023qdiffusion, shang2023ptqdm, he2023ptqd, huang2024tfmqdm} have employed it for small-scale diffusion models, achieving nearly lossless compression at 4-8 bits. 
However, for higher compression rates, a more sophisticated representation might be necessary to maintain the original model's performance.

Vector quantization (VQ) is a promising candidate due to its flexibility and ability to account for the non-uniformity of data distribution, thus offering smaller quantization errors for the same compression rate.
The encoded vectors are stored as low-bit integer indices, \emph{codes}, which point to vector representations, or \emph{codewords}, in the corresponding codebooks shared within a model layer. 
Various VQ extensions~\citep{aq, rvq, pq, lsq, lsq++} employ multiple codebooks and hence provide even better compression quality.
Recently, vector quantization has demonstrated state-of-the-art results in low-bit compression of large language models~\citep{egiazarian2024extreme, tseng2024quip}. 
Therefore, it is appealing to explore the application of vector quantization for large-scale diffusion models.

\paragraph{Contributions.} 
This work presents a novel post-training quantization method for large-scale text-to-image diffusion models.
Our PTQ method relies on \textit{additive quantization} (AQ)~\citep{aq}, the generalized multi-codebook vector quantization (MCQ) formulation originated in vector search for extreme vector compression.
We explore AQ specifically for billion-parameter diffusion U-Net architectures~\citep{podell2024sdxl, sauer2023adversarial} and propose several practical techniques for their fast and accurate compression.
Notably, capitalizing on the increased capacity of AQ codebooks compared to uniform quantization, we demonstrate that the initial calibration can largely benefit from additional codebook tuning. 
Following~\citep{he2024efficientdm}, we propose to fine-tune the quantized model to mimic the final U-Net predictions of the full-precision model.
Unlike typical settings in quantization-aware training (QAT)~\citep{Nagel2022OvercomingOI, Esser2020learned}, the fine-tuning approach does not require training on large-scale datasets and takes a tiny fraction of the time needed to train the original model. 
We demonstrate that fine-tuning significantly eliminates the discrepancy with the full-precision model, unlocking highly accurate $3$-bit weight compression for PTQ of billion-scale diffusion models.
To sum up, our contributions can be formulated as follows:

\begin{itemize}
    \item We explore vector-based PTQ strategies for text-to-image diffusion models and demonstrate that the compressed models yield higher quality text-to-image generation than the scalar alternatives under the same bit-widths. 
    Furthermore, we describe an effective fine-tuning technique that further closes the gap between the full-precision and compressed models, leveraging the flexibility of the vector quantized representation. 
    
    \item To illustrate the power of our technique, we compress the weights of SDXL~\citep{podell2024sdxl}, the state-of-the-art text-to-image diffusion model with 2.6B parameters, down to $3$ bits per parameter. 
    Extensive human evaluation and automated metrics confirm the superiority of our approach over previous diffusion compression methods under the same bit-widths. 

    \item We illustrate that our approach can be effectively applied to distilled diffusion models, such as SDXL-Turbo~\citep{sauer2023adversarial}, achieving nearly lossless 4-bit compression. 
    This highlights the potential for combining integrating text-to-image model quantization with other diffusion acceleration techniques. %

\end{itemize}

\vspace{-12pt}
\section{Related work}
\vspace{-5pt}
\paragraph{Efficient diffusion models.}
Text-to-image diffusion models~\citep{saharia2022photorealistic, podell2024sdxl, esser2024scaling, rombach2021highresolution, pernias2023wuerstchen, betker2023improving} are a class of generative models that gradually transform noise samples into realistic images corresponding to textual prompts.
One of the most prominent challenges in diffusion modeling is a sequential sampling procedure resulting in significantly slower inference compared to feed-forward alternatives~\citep{goodfellow2014generative, kingma2022autoencoding, kang2023gigagan, Sauer2023ARXIV}. 

Several major research directions address the efficiency of text-to-image diffusion models.
Diffusion distillation~\citep{sauer2024fast, sauer2023adversarial, luo2023latent, song2023consistency, meng2023distillation} and advanced sampling algorithms~\citep{zhou2024fast, lu2022dpm, zhao2023unipc} aim to reduce the number of sampling steps in the sequential inference of diffusion models. 
Intermediate activation caching~\citep{ma2023deepcache, wimbauer2024cache, li2023faster}, efficient architecture designs~\citep{rombach2021highresolution, podell2024sdxl} and structured pruning~\citep{kim2023bksdm, fang2023structural} focus on faster sampling at the architecture level.
Model quantization methods~\citep{li2023qdiffusion, shang2023ptqdm, he2023ptqd, huang2024tfmqdm, Chang2023EffectiveQF, wang2024quest} aim to reduce the memory footprint and runtime by compressing the weights and/or activations of diffusion models to compact representations.

\paragraph{Model quantization.}
This work focuses on model quantization, which has shown remarkable accuracy-compression trade-offs in the context of LLM compression~\citep{dettmers2023spqr, chee2023quip, egiazarian2024extreme, tseng2024quip}.
The quantization methods can generally be categorized into two classes: \emph{post-training quantization} (PTQ)~\citep{li2021brecq, nagel20a} and \emph{quantization-aware training} (QAT)~\citep{Nagel2022OvercomingOI, Esser2020learned}. 
PTQ performs quantization and weight adjustment on top of the pretrained model and typically requires largely lower costs compared to the training process. %
QAT methods, in contrast, undergo multiple rounds of training and quantization iterations.
While they are likely to achieve higher accuracy than PTQ, they require more sophisticated and resource-intensive quantization procedures.

In practice, quantization is applied to linear or convolution layers, consuming the prevalent portion of the model weights in modern deep learning architectures~\citep{frantar2022obc,10.5555/3524938.3525851, khan2020survey}.
Uniform scalar quantization first scales full-precision weights to the specified value range and then rounds them into low-bit fixed-point values.
Although uniform quantization is efficient and simple to use, it disregards non-uniformity of the underlying distributions and presence of outliers~\citep{dettmers2023spqr, lin2024awq, chee2023quip}, leading to large quantization errors at low-bit compression.
In contrast, non-uniform scalar quantization assigns individual weights
to the scalar values $c_i$, \emph{codewords}, from a codebook containing $k{=}{2^B}$ elements $C{=}\{c_1, ..., c_k\}$.
Each weight is associated with a $B$-bit index $i$, \emph{code}, of the corresponding codeword $c_i$.
The codebooks are more flexible in capturing the underlying weight distributions than uniform grids, making non-uniform quantization more accurate under the same bit-width. 
Nevertheless, uniform quantization remains popular due to its highly efficient encoding and decoding processes.

Vector quantization (VQ) \citep{vq1, vq2} extends non-uniform scalar quantization by increasing the dimensionality of the codewords. 
Thus, the VQ-based methods approximate a group of consecutive weights by a vector from the codebook.
In our work, we exploit multi-codebook generalization of VQ, \emph{additive quantization} (AQ)~\citep{aq, lsq, lsq++}. 
AQ represents the weight groups as a sum of $M$ codewords chosen from multiple learned codebooks $C_1, ... C_M$. 
Weight group size, the number of codebooks $M$, and codebook size $k=2^B$ dictate the trade-off between memory footprint and compression accuracy. 

After initialization, the codebooks and other quantization parameters, e.g., scaling factors, are \emph{calibrated} to approximate the activation distribution of the original model. 
The activation statistics are collected by running the model on small representative \emph{calibration data}. 
Recent PTQ calibration strategies~\citep{nagel20a, frantar2022obc, Frantar2023OPTQ, egiazarian2024extreme} simplify the model quantization problem to layer-wise quantization, enabling scalability to billion-parameter models. 
These methods minimize the MSE between the outputs of the full-precision weights $W \in \mathbb{R}^{d_{out}{\times}d_{in}}$ and the corresponding quantized weights $W_q$ given calibration inputs $X$ for each layer individually:
\begin{equation}
\label{eq:basic_objective}
\vspace{-2px}
\underset{W_q}{\arg\min} || W X - W_q X ||^2_2
\vspace{-2px}
\end{equation}
For AQ quantized $W_q$, the optimal configuration is learned via alternating optimization of codes and codebooks. 
Our method adopts the learning procedure in~\citet{egiazarian2024extreme}, where AQ is applied to LLM compression.

\paragraph{Quantization of diffusion models.}
Previous diffusion PTQ approaches~\citep{li2023qdiffusion, shang2023ptqdm, he2023ptqd, huang2024tfmqdm} employ uniform scalar quantization and calibrate the quantization parameters using adaptive rounding~\citep{nagel20a} or simple min-max initialization~\citep{nagel2021white}.
Some studies introduce diffusion-specific calibration data collection methods~\citep{li2023qdiffusion, shang2023ptqdm} and investigate the importance of different diffusion time steps and model layers~\citep{li2023qdiffusion, shang2023ptqdm, yang2023efficient, Chang2023EffectiveQF, huang2024tfmqdm}.
Other works address the accumulated quantization error caused by sequential diffusion sampling with quantized models~\citep{he2023ptqd, li2023qdm}.
A few propose using variable bit widths for activations at different time steps~\citep{he2023ptqd, tang2023posttraining}.

Recent works~\citep{he2024efficientdm, wang2024quest} have demonstrated that effective quantization-aware fine-tuning (QAFT) may achieve state-of-the-art quantization performance on small-scale benchmarks.
EfficientDM~\citep{he2024efficientdm} introduces quantization-aware parameter-efficient fine-tuning for diffusion models and runs full-precision model distillation after initial calibration.
TQD~\citep{so2023temporal} fine-tunes diffusion models in a QAT manner and
employs additional MLP modules to estimate quantization scales for each step.
QuEST~\citep{wang2024quest} proposes to fine-tune only the most vulnerable to quantization parts of the diffusion model: time-embedding, attention-related quantized weights, and activation quantizer parameters.

Contrary to the approaches mentioned above, we investigate more versatile weight quantizers for large-scale diffusion model compression. 
In this work, we focus solely on weight quantization and leave the investigation of activation quantization for future work.

\begin{figure*}[t!]
    \vspace{-8mm}
    \hspace{-2mm}\includegraphics[width=1.06\linewidth]{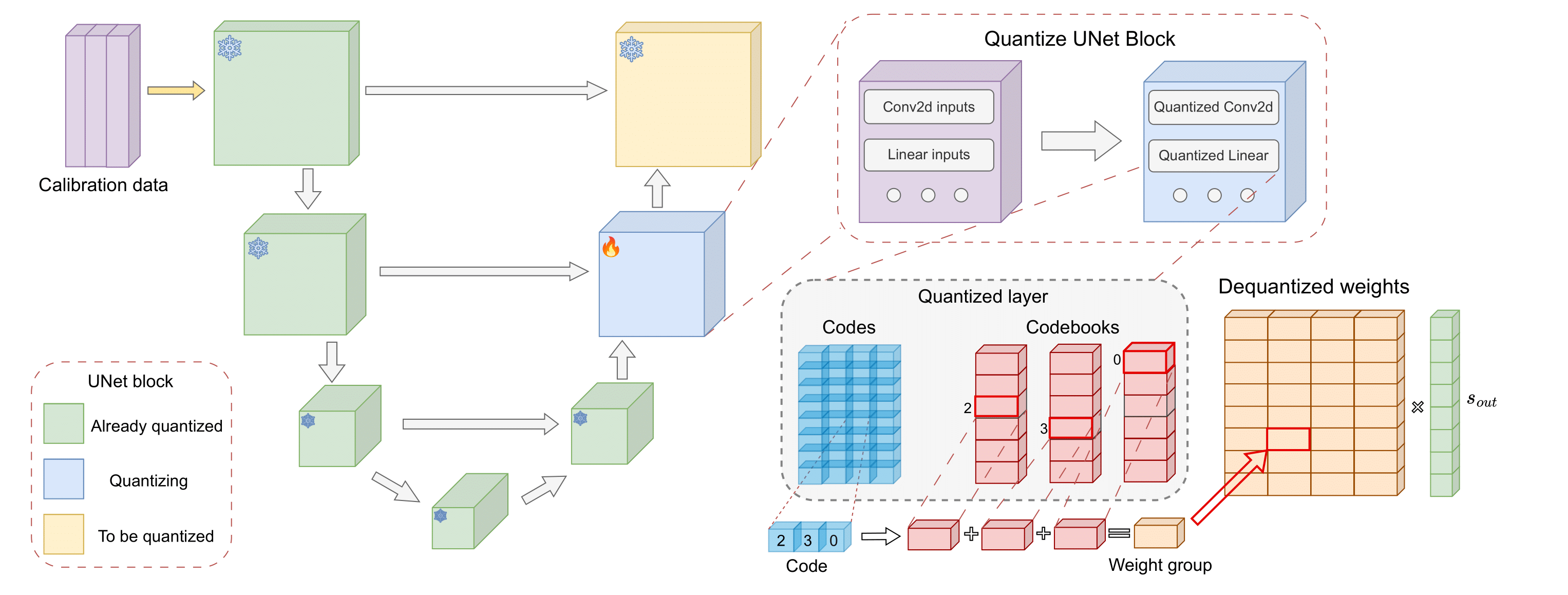}
    \vspace{-20pt}
    \caption{Overview of the proposed layer-wise calibration procedure before fine-tuning.}
    \vspace{-5pt}
    \label{fig:overview}
\end{figure*} 

\vspace{-10pt}
\section{Method}
\vspace{-6pt}
In this section, we present a VQ-based method adapted for text-to-image diffusion model compression.
Section~\ref{sect:method_arch} introduces the strategy for applying the existing VQ algorithms to the U-Net architecture, commonly used in diffusion models. 
Then, in Section~\ref{sect:method_training}, we describe the proposed calibration method comprising of the layer-wise calibration and global fine-tuning procedures.
In Section~\ref{sect:inference_method}, we discuss the inference procedure for the VQ-compressed diffusion models.

\vspace{-5pt}
\subsection{Vector Quantization of Text-to-Image Models}\label{sect:method_arch}
\vspace{-3pt}

Existing vector quantization algorithms are designed for linear models
used in retrieval applications \citep{pq, aq, lsq, lsq++} or large language models with extremely large hidden dimension~\citep{egiazarian2024extreme,tseng2024quip}.
Naive transfer of the method introduced in ~\citet{egiazarian2024extreme, vanbaalen2024gptvq} would not allow one to achieve practically useful compression rates due to significant architectural differences between LLMs and text-to-image diffusion models.

Specifically, unlike LLMs, which are built from homogeneous transformer blocks with the same number of hidden features, modern text-to-image diffusion models contain several heterogeneous components, including the diffusion model itself -- typically a U-Net~\citep{ronneberger2015u} or 2D image Transformer~\citep{esser2024scaling, Peebles2022DiT}, one or more text encoders, autoencoders for latent diffusion models~\citep{rombach2021highresolution} or super-resolution models for cascaded diffusion models~\citep{saharia2022photorealistic}. 

As an illustrative example, we consider Stable Diffusion XL (SDXL)~\citep{podell2024sdxl}: a latent diffusion model consisting of a large 2D U-Net diffusion model with ${\sim}2.6$B parameters, two text encoders, a variational autoencoder~\citep{kingma2022autoencoding}, and an optional refiner network.
Text encoders are regular transformer models with known high-performance quantization algorithms, including vector quantization~\citep{egiazarian2024extreme, tseng2024quip, vanbaalen2024gptvq}. 
The variational encoder is a relatively small network with less than 0.1B parameters.
Since the overall amount of computation and inference time is dominated by the iterative application of the diffusion model, only the compression of the U-Net model is of practical interest. 

The U-Net backbone adopted in SDXL comprises stacks of residual and transformer blocks at different resolutions. 
The standard U-Net architecture includes an encoder, middle block, and decoder. 
The number of channels increases with the decrease in the feature map size after pooling operations. 
Specifically, SDXL U-Net has 3 downsampling levels, with $320$ channels on the top level and the multiplication rates $[1, 2, 4]$ at different resolutions. 
A typical number of channels in a layer is in the order of $10^{2}{-}10^{3}$, which is smaller by an order of magnitude than modern LLMs. 
Notably, the 16-bit codebooks with a group size of $8$, that yield the highest quality in low-bit LLM compression~\citep{egiazarian2024extreme}, occupy $2^{16} \cdot 8 \cdot 2 = 2^{20}$ bytes and exceed the size of many SDXL layers, making such quantization configuration impractical. 
Thus, one has to work with small codebooks (of size $2^6$, $2^{8}$), and capacity could be increased by increasing the number of codebooks. 

Another aspect is the presence of convolutional layers in addition to linear in ResNet-like~\citep{he2016residual} blocks. 
Following the common practice~\citep{DBLP:journals/corr/abs-2106-08295} for CNN quantization, we group weights along the input channel dimension rather than the kernel spatial dimensions.

We quantize all convolutional and linear layers within the U-Net blocks, with a few exceptions:
\vspace{-5pt}
\begin{itemize}
    \item We do not quantize the first and last convolutional layers: the first layer has an input dimension of $4$, and the last layer has a commensurate output dimension. These layers constitute a tiny fraction of an overall number of parameters, and there is no benefit from their compression.
    \item Time embedding layers are not quantized inspired by recent works~\citep{huang2024tfmqdm, wang2024quest} demonstrating the importance of the temporal features for quantized diffusion models.
    In addition, for a given input sample, timestep projections accept only a single vector instead of a whole feature map, as most linear projections and convolutional layers do. 
    Finally, they incur small additional memory overhead. 
    Therefore, we do not quantize these layers based on practical considerations. 
\end{itemize}

\vspace{-8pt}
\subsection{Calibrating Vector-Quantized Diffusion Models}
\vspace{-3pt}
\label{sect:method_training}

Here, we discuss the calibration procedure used to determine the optimal quantized weight configuration. 
The procedure involves two stages: 
i) layer-wise calibration and 
ii) global fine-tuning.
Below, we discuss both stages in more detail.

\paragraph{Layer-wise calibration.}
First, the calibration set is collected by running the diffusion sampling on a small set of calibration prompts.
Text-to-image diffusion models alter intermediate generation steps when using classifier-free guidance (CFG)~\citep{ho2022classifier}. 
Therefore, we run the diffusion sampling with a default CFG scale $5$ and considering both unconditional and conditional inputs as separate calibration samples. 
Unlike some prior works~\citep{shang2023ptqdm, li2023qdiffusion}, we collect calibration data for all diffusion timesteps and sample them uniformly.

The calibration procedure is performed sequentially for predefined consequent layer subsets.
Specifically, the model, partially quantized up to the $i{-}$th subset, collects the activations for all layers within the $i{+}1{-}$th subset. 
Then, these layers are quantized in parallel, and the procedure goes to the following subset.
The number of layers in subsets controls the calibration speed, memory consumption, and overall model quality.
The quantization of all layers is the fastest, but storing the activation statistics requires a large amount of memory. 
In addition, using the original model activations to quantize deeper layers does not account for the output changes in the previous layers. 
Another extreme is to recollect the data after each quantized layer sequentially. 
However, the number of layers in modern diffusion models is several hundred, making such an option intolerably slow. 
Therefore, we choose the entire stack of blocks (both convolutional and transformer) for a given resolution as a subset of weights quantized at once. 
Since the SDXL U-Net has 3 downsampling and upsampling levels and a middle block, this constitutes 7 subsets in total.
This choice provides a preferable trade-off between accuracy, memory efficiency, and calibration runtime.

Also, previous diffusion PTQ methods~\citep{li2023qdiffusion, he2023ptqd, shang2023ptqdm} collect raw model activations $X \in \mathbb{R}^{n{\times}d_{in}}$, where $n$ is the calibration dataset size multiplied by the spatial dimension $h_l{\times}w_l$ for a layer $l$.
We notice that this limits their scalability to large text-to-image models and calibration sets. 
Instead, following the practices in LLM quantization~\citep{frantar-gptq, egiazarian2024extreme, vanbaalen2024gptvq},
we collect $XX^T \in \mathbb{R}^{d_{in}{\times}d_{in}}$ and modify the objective~(\ref{eq:basic_objective}) as follows: $||W X - W_q X||_2^2 = ||(W - W_q) X||_2^2 = \left\langle (W - W_q) XX^T, (W -W_q)\right\rangle_F$.
In our experiments, this vastly reduces the storage of calibration data.  

Finally, we closely adopt the codebook learning approach from~\citet{egiazarian2024extreme}.
In addition to the codebooks, this method employs learnable scales $s_{out} \in \mathbb{R}^{d_{out}}$ to multiply each output dimension of the dequantized matrix $W_q \in \mathbb{R}^{d_{out}{\times}d_{in}}$.
Thus, the dequantized matrix can be represented as $s_{out} W_q$.
\Cref{fig:overview} illustrates the proposed layer-wise calibration procedure.

\paragraph{Global fine-tuning.}
Since fine-tuning is known to boost the performance of vector-quantized models~\citep{egiazarian2024extreme, tseng2024quip},
we equip our calibration procedure with an end-to-end fine-tuning stage. 
Previous diffusion quantization approaches~\citep{li2023qdiffusion, shang2023ptqdm, huang2024tfmqdm} perform block-wise fine-tuning for each residual or attention block right after their layer-wise calibration.
However, the block-wise fine-tuning is unaware of the final model prediction.
Therefore, we employ a more accurate solution by minimizing MSE between global U-Net predictions at each denoising timestep. 
We notice that there is no need for fine-tuning on large-scale data; 
a few thousand calibration samples generated with the full-precision U-Net are sufficient.
In fact, such fine-tuning is an instance of model distillation~\citep{hinton2015distilling} and helps to compensate inter-layer errors caused by independent layer-wise or block-wise calibration. 

The model's forward pass is differentiable with respect to the codebook vectors; therefore, these can be optimized like any other parameter inside the U-Net network.
In addition, all non-quantized layers are made trainable to compensate the quantization error. 

The resulting calibration procedure is described in Appendix~\ref{app:alg} in Algorithms~\ref{alg:quantization},~\ref{alg:finetuning}.
First, the algorithm iterates over the U-Net down, middle, and up blocks and accumulates input activations within each block by running the diffusion sampling loop with classifier-free guidance~\citep{ho2022classifier}. 
Then, the layers within each block are quantized and calibrated with the algorithm inherited from AQLM~\citep{egiazarian2024extreme}.
Once the model is quantized, we fine-tune the entire model to mimic the teacher output. 
In the following, we denote the entire calibration method as VQDM.
\vspace{-7pt}
\subsection{Inference procedure}
\vspace{-5pt}
\label{sect:inference_method}

Uniform scalar quantization facilitates faster inference by using more efficient low-precision calculations while maintaining a constant number of arithmetic operations.
In contrast, the VQ inference procedure first precomputes \textit{look-up tables} (LUTs) w.r.t. the learned codebooks and then performs matrix multiplication by retrieving values from the LUTs and summing them to obtain the final result.
This technique was originally proposed for quantized nearest neighbor search~\citep{pq,aq} and can be adapted to quantized weight matrices~\citep{blalock2021multiplying,egiazarian2024extreme}.
Though such a procedure involves fewer arithmetic operations, we observe that it significantly slows down the diffusion inference on high-end GPUs and CPUs due to the absence of specific hardware realizations for efficient look-ups. 
In other words, software look-ups are slower than hardware-optimized multiplications with tensor cores or AVX-512.

As a result, we found it more effective to simply dequantize the weights into half- or full-precision values in runtime and perform standard matrix multiplication that benefits from hardware optimizations.
Still, the proposed quantization method carries additional computational overheads caused by the dequantization procedure during inference.
In~\Cref{sec:inference}, we measure the runtime overheads for our CPU and GPU implementations.
We believe that this overhead is not a fundamental property of vector quantization but a quirk of modern hardware. 
Several recent studies~\citep{abouelhamayed2023pqa,zhu2024scalable} explore this further and achieve highly efficient vector quantization inference using FPGA (Field Programmable Gate Arrays) hardware. 
Both studies use Intel Agilex FPGAs.

\vspace{-8pt}
\section{Experiments}
\vspace{-5pt}

\label{experiments}
\begin{figure*}[t!]
    \vspace{-5mm}
    \includegraphics[width=0.98\linewidth]{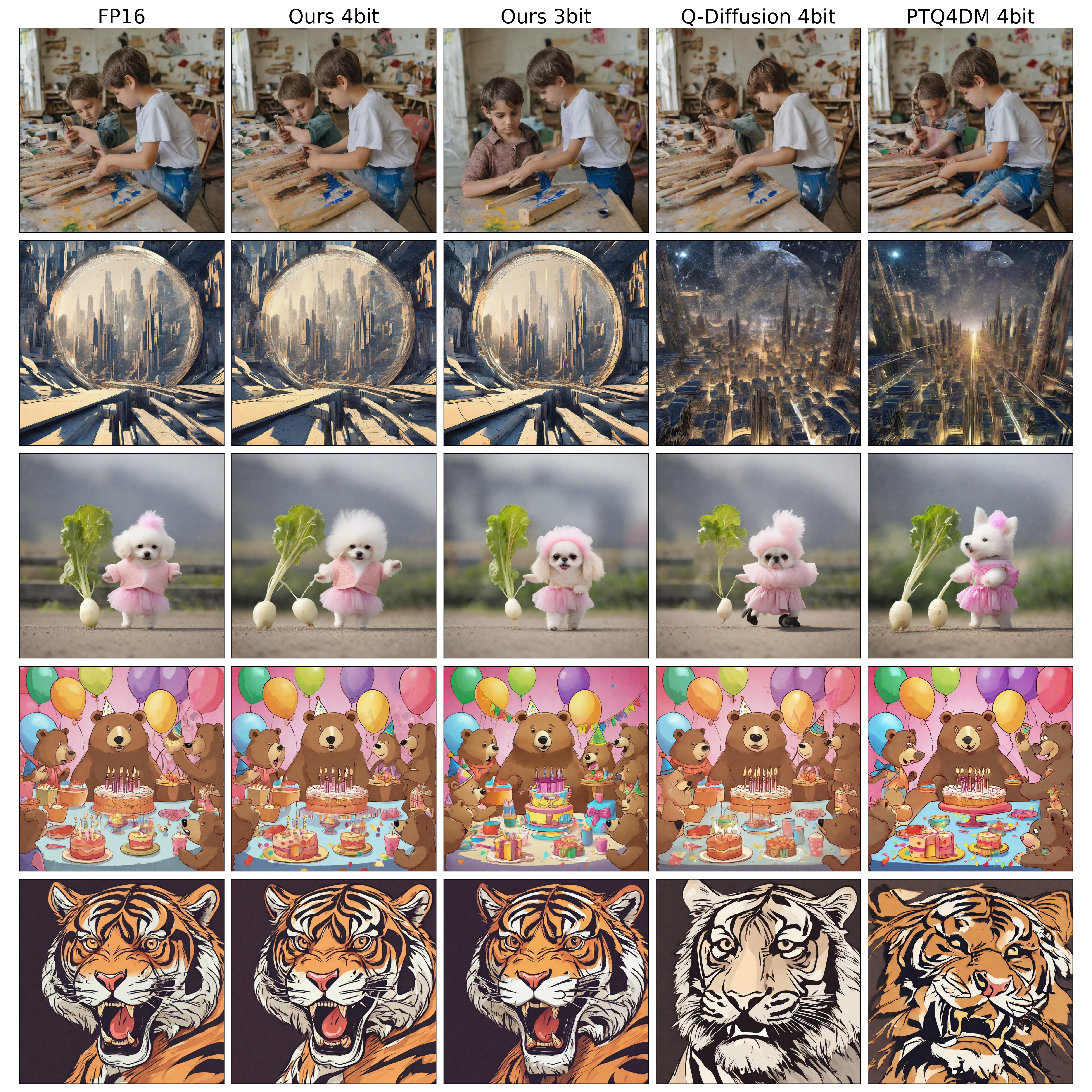}
    \vspace{-2mm}
    \caption{Qualitative comparison of SDXL compressed with VQDM and the baselines.}
    \label{fig:qualitative_overview}
    \vspace{-10pt}
\end{figure*} 

We begin by comparing VQ-compressed SDXL to the full-precision model and previous diffusion quantization methods.
Then, we measure the inference overheads and realised memory reduction for our GPU and CPU implementations.
Next, we conduct an ablation study on the design choices discussed in our method. 
Finally, we apply VQDM to the distilled diffusion model, SDXL-Turbo.%
\vspace{-7pt}
\subsection{Experimental setup}
\vspace{-5pt}
\begin{table*}[t!]
\vspace{-0.5em}
\footnotesize
\centering
\scriptsize
\setlength\tabcolsep{2.47pt}
\renewcommand{\arraystretch}{1.2}
\begin{tabular}{lcc|ccc}
 \toprule
  \bf{Model} & \bf{Method} & \bf{Avg bits} & \bf{Pickscore$\uparrow$} & \bf{CLIP$\uparrow$} & \bf{FID$\downarrow$}\\
  \midrule
 \rowcolor[HTML]{dddddd}
 \multirow{5}{*}{SDXL}  & Original model & 32 & 0.226 & 0.357 & 18.99 \\
  & VQDM & 4.15 & \bf{0.226} & \bf{0.356} & \bf{19.11} \\
  & VQDM & 3.15 & 0.225 & 0.355 & 19.18 \\
  & VQDM & 2.15 & 0.219 & 0.341 & 22.14 \\
  & Q-Diffusion & 4 & 0.225 & 0.355 & 19.30 \\
  & PTQ4DM & 4 & 0.224 & 0.353 & 19.78 \\	
 \bottomrule
\end{tabular}
\vspace{-3pt}
\caption{Evaluation of quantized \textsc{SDXL} models for different bit-widths in terms of automatic metrics.}
\vspace{-15pt}
\label{tab:SDXLmetrics}
\end{table*}

As a primary quality measure, we consider side-by-side human evaluation (SbS) on $128$ prompts specifically selected from PartiPrompts~\citep{yu2022scaling}, following~\citet{sauer2024fast}.
We generate 2 images for each prompt and report the SbS score as a portion of the student wins plus half of the tied evaluations.
To our knowledge, this is the first work that conducts the human study to assess quantized diffusion models.
More details about the human evaluation setup are in Appendix~\ref{app:sbs}.

Additionally, we employ the standard metrics such as FID~\citep{fid} and CLIP Score~\citep{clip} and also evaluate PickScore~\citep{kirstain2023pickapic}, designed to estimate the human preference score. 
The automated metrics are calculated on $5000$ prompts from the COCO2014 validation set~\citep{lin2014microsoft}.

As a compression measure, we report an average number of bits per model weight in all convolutional and linear layers, including the unquantized ones.
The codebook sizes are also included, divided by the overall number of weights. 

For VQDM, if not otherwise stated, we set the group size and number of bits per codebook to $8$ and vary the number of codebooks from $2$ to $4$ to achieve $2{-}4$ bit compression, respectively.
\vspace{-5pt}
\subsection{Comparison with baseline methods}
\vspace{-2pt}
Here, we evaluate our approach on one of the most widely adopted open source text-to-image models, SDXL, containing $2.6$B parameters and compare VQDM to previous PTQ approaches.

\paragraph{Baselines.}
We select Q-Diffusion~\citep{li2023qdiffusion} as our primary baseline from the family of uniform scalar PTQ techniques. 
The quantization is applied to all linear and convolutional layers in the U-Net model, followed by a calibration process. 
To form the calibration set, Q-Diffusion splits sampling time steps into equal intervals and uses one time step from each interval for each of the $n$ prompts.
Additionally, Q-Diffusion proposes ``shortcut-splitting'' in up-sample ResNet blocks of U-Net to mitigate their abnormal activation distribution.

Changing the strategy for collecting timesteps for calibration to sampling from a normal distribution skewed to $t=T_0$ and removing ``shortcut-splitting'' results in PTQ4DM method~\citep{shang2023ptqdm}, another baseline we compare with.
In contrast to these methods, we use all timesteps for each calibration prompt. 
Another difference is that we do not quantize time embedding layers.

While EfficientDM~\citep{he2024efficientdm} and QuEST~\citep{wang2024quest} are also competitive baselines, adapting their code to text-to-image generation and the SDXL model requires significant effort.
Therefore, we left the comparison with these methods for future work.

\paragraph{Experimental results.}
First, we compare our method with the uncompressed base model.
\Cref{tab:SDXLmetrics} reveals that VQDM allows to efficiently compress SDXL to $4.15$ bits per parameter with a minor drop in visual quality and prompt alignment.
According to the human study, the samples generated with the $4.15$ bit VQDM model were almost indistinguishable for human annotators, resulting in an SbS score of $44.5\%$ with a p-value $>0.1$ for the null hypothesis claiming that both models perform equally.
The $3.15$ bit model also demonstrates highly promising performance, being marginally worse than the full precision model in terms of SbS score.

Then, we adapt the Q-Diffusion code for the SDXL U-Net architecture and calibrate the quantized U-Net using 4 bits for weight quantization and keeping activations in full precision to match our setup. 
We use $n=64$ calibration prompts and run calibration for 10000 iterations for each layer and block.
These hyperparameters were tuned to match the training time with our method without compromising the quality of the baselines.

\begin{figure}
\vspace{-5pt}
\centering
\begin{subfigure}{.5\textwidth}
  \centering
  \includegraphics[width=\linewidth]{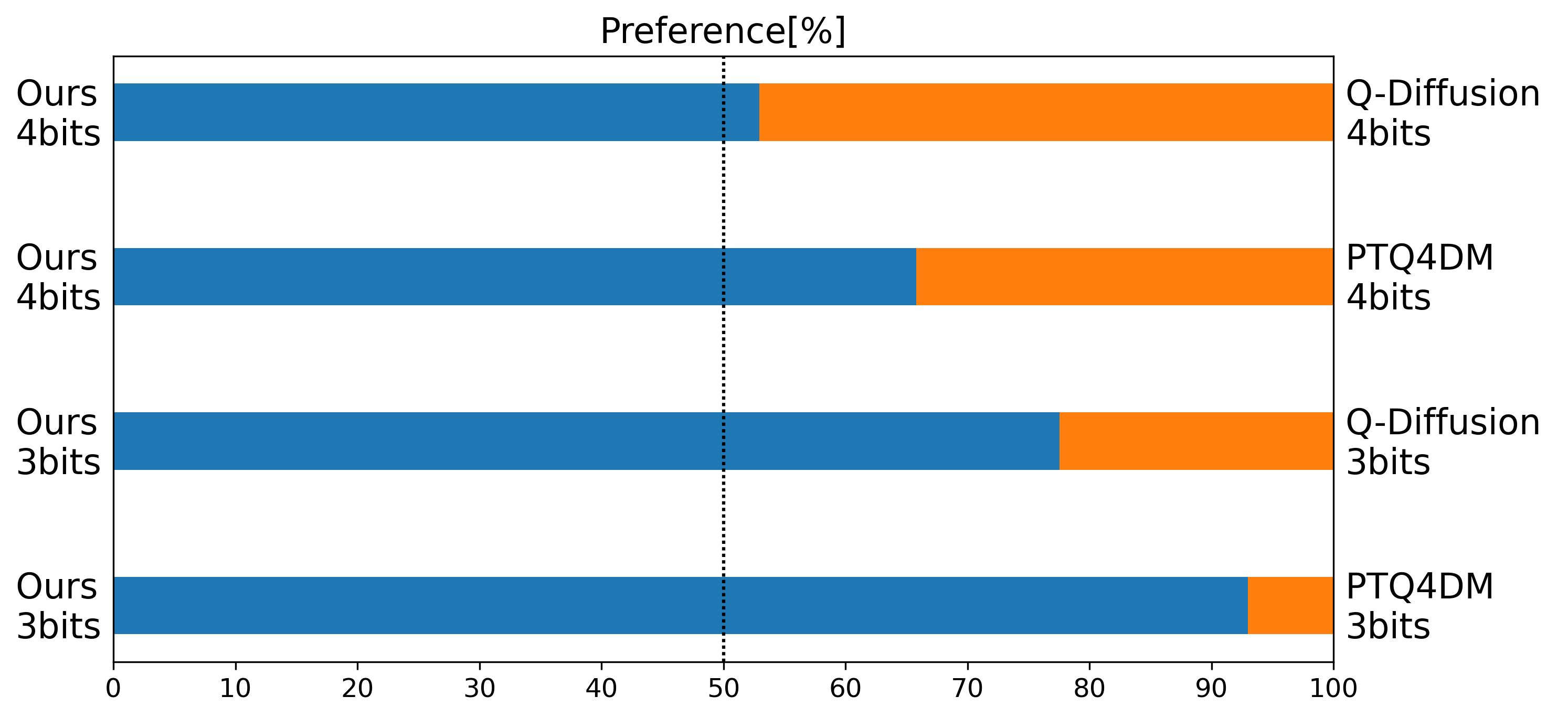}
  \label{fig:ours_comp_sbs}
\end{subfigure}%
\begin{subfigure}{.5\textwidth}
  \centering
  \includegraphics[width=\linewidth]{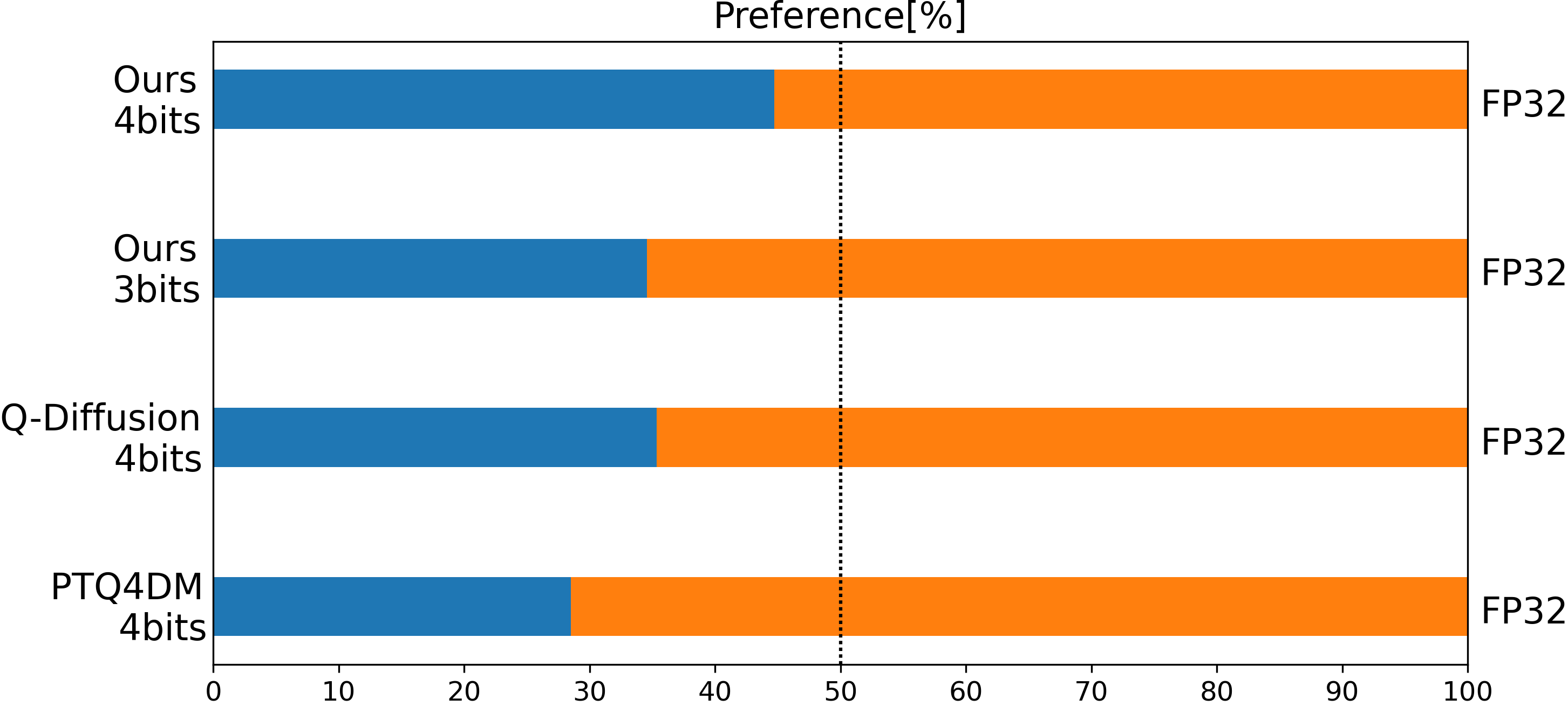}
  \label{fig:ours_sdxl_sbs}
\end{subfigure}
\vspace{-25pt}
\caption{\textbf{Human preference study.} 
\textbf{Left.} Comparison between VQDM and the baselines.
\textbf{Right.} Comparison between the quantized and full-precision models.}
\label{fig:test}
\vspace{-10pt}
\end{figure}

In terms of automated metrics, the results in~\Cref{tab:SDXLmetrics} indicate that 4 bit VQDM outperform both 4-bit baselines while 3-bit matches their performance.
Human evaluation in~\Cref{fig:test} and qualitative comparison in~\Cref{fig:qualitative_overview} support this claim. 

\vspace{-8pt}
\subsection{Inference performance}
\vspace{-3pt}
\label{sec:inference}
In~\Cref{tab:cpu_gpu_inference}, we report the dequantization runtime overheads and released memory reduction for the SDXL model using our GPU and CPU implementations supporting VQDM.

For GPU measurements, we perform half-precision SDXL inference and use batch size $8$ to saturate GPU utilization. 
The experiments are run on a single NVIDIA A100 80Gb. 
Compared to the FP16 model, VQDM allows up-to $5{\times}$ memory reduction at the cost of ${\sim}50\%$ runtime overhead. 

For CPU measurements, we run full-precision SDXL inference since CPUs generally do not support effective operations in half-precision. 
The experiments are run on a single core of an Intel Sapphire Rapids CPU.
We observe that memory reductions of up to $9.7{\times}$ can be realized in practice with relatively low runtime overheads ($23{-}26\%$).

\begin{figure*}[t]
    \centering
    \includegraphics[width=1.0\linewidth]{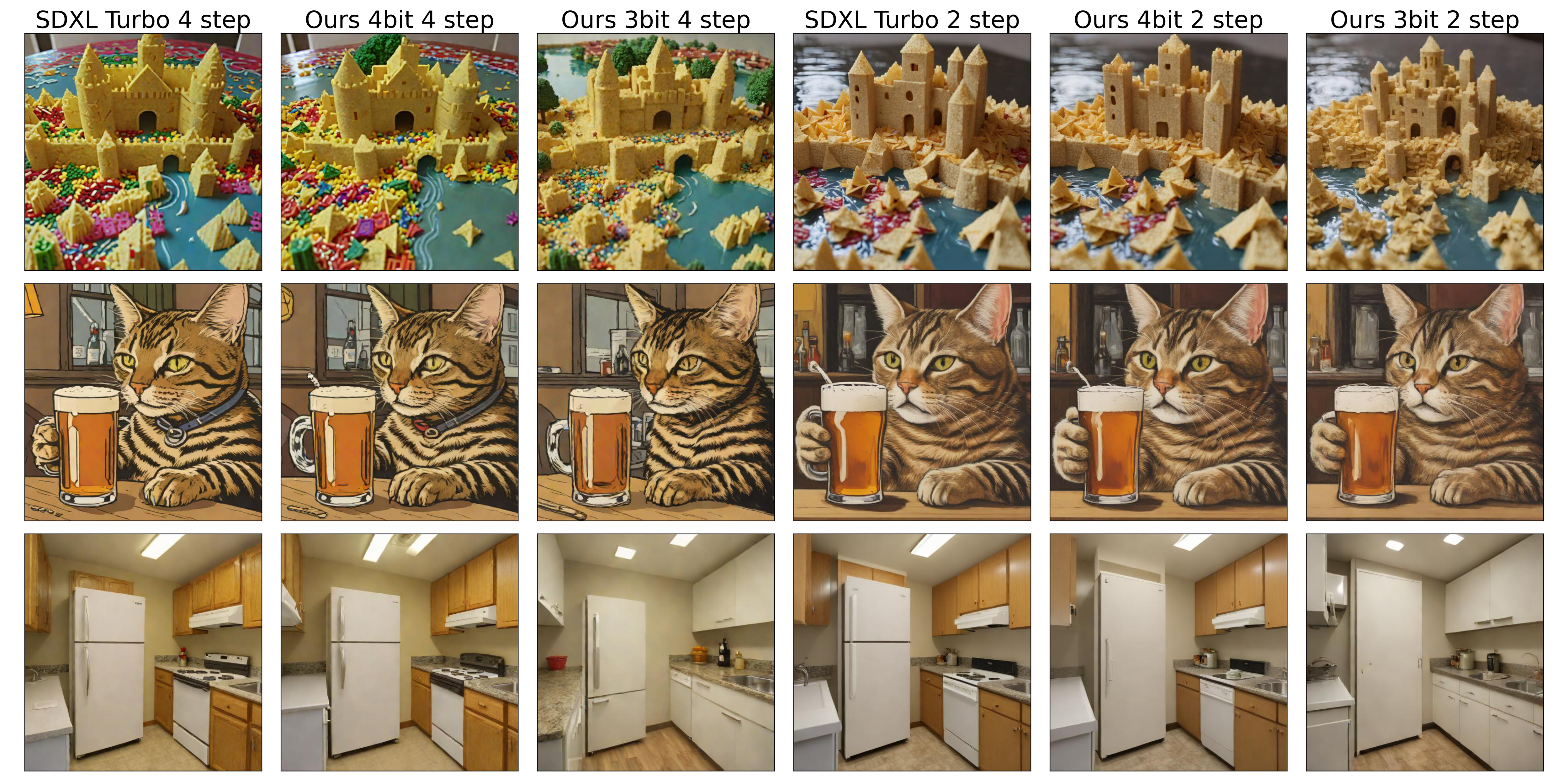}
    \vspace{-15pt}
    \caption{Qualitative comparison of SDXL-Turbo quantized with VQDM and the full-precision model for different sampling steps.}
\vspace{-10pt}
\end{figure*} 

\begin{table}[h]
\centering
\resizebox{0.82\textwidth}{!}{%
\begin{tabular}{c|cc|cc}
\toprule
& \multicolumn{2}{c|}{GPU implementation FP16} & \multicolumn{2}{c}{CPU implementation FP32} \\
\midrule
\bf{Avg. bits} & \bf{Runtime overhead} & \bf{Memory reduction} & \bf{Runtime overhead} & \bf{Memory reduction} \\
\midrule
4.15 & $49\%$ & $3.76{\times}$ & $26\%$ & $7.42{\times}$ \\
3.15 & $48\%$ & $5.06{\times}$ & $23\%$ & $9.70{\times}$ \\
\bottomrule
\end{tabular}}
\vspace{1mm}
\caption{Inference results for our GPU and CPU implementations, showing decompression overheads and released memory savings for 3-4-bit VQDM configurations.}
\vspace{-5pt}
\label{tab:cpu_gpu_inference}
\end{table}

\vspace{-5pt}
\subsection{Ablation}
\vspace{-3pt}
\label{sec:ablation}

\begin{table}[]
\begin{minipage}[b]{0.48\linewidth}
\centering
\scriptsize
\setlength\tabcolsep{2.47pt}
\renewcommand{\arraystretch}{1.2}
\begin{tabular}{lc|cccc}
 \toprule
  \bf{Method} & \bf{Avg bits} & \bf{Pickscore$\uparrow$} & \bf{CLIP$\uparrow$} & \bf{FID$\downarrow$} & \bf{SbS,\% $\uparrow$}\\
 \midrule
\multirow{2}{*}{VQDM-all} & 4.06 & 0.226 & 0.357 & 18.88 &  43.2 \\
 & 3.05 & 0.225	& 0.355	& 19.30 & 31.1 \\
  \midrule
\multirow{2}{*}{VQDM} & 4.15 & 0.226 & 0.356 & 19.11 &  44.7 \\
 & 3.15 & 0.225 & 0.355 & 19.18 & 34.6 \\
 \bottomrule
\end{tabular}
\vspace{1mm}
\caption{
Comparison of VQDM with and without quantization of the timestep embedding layers. VQDM-all denotes quantizing the timestep embedding layers.
}
\label{tab:time_embedding_ablation}

\end{minipage}
\hspace{3mm}
\begin{minipage}[b]{0.48\linewidth}
\centering
\scriptsize
\setlength\tabcolsep{2.47pt}
\renewcommand{\arraystretch}{1.2}
\begin{tabular}{lc|cccc}
 \toprule
  \bf{Method} & \bf{Avg bits} & \bf{Pickscore$\uparrow$} & \bf{CLIP$\uparrow$} & \bf{FID$\downarrow$} & \bf{SbS,\% $\uparrow$}\\
  \midrule	
 \rowcolor[HTML]{dddddd} Original model & 32 & 0.226 & 0.357 & 18.99 & 50.0 \\
 \midrule
\multirow{3}{*}{VQDM w/ FT} & 4.15 & 0.226 & 0.356 & 19.11 & 44.7 \\
 & 3.15 & 0.225 & 0.355 & 19.18 & 34.6 \\
 & 2.15 & 0.219 & 0.341 & 22.14 & 9.2 \\
  \midrule
\multirow{3}{*}{VQDM w/o FT} & 4.15 & 0.225 & 0.355 & 20.24 &  34.2 \\
 & 3.15 & 0.221 & 0.347 & 24.88 & -- \\
 & 2.15 & 0.206 & 0.287 & 86.88 & -- \\
 \bottomrule
\end{tabular}
\vspace{-1mm}
\caption{
Comparison of VQDM quantization settings before and after fine-tuning (FT).}
\label{tab:ft_ablation}
\vspace{-5pt}

\end{minipage}
\vspace{-5pt}
\end{table}

Next, we investigate the influence of the fine-tuning and architecture choices for VQDM on final model performance.
The results in~\Cref{tab:ft_ablation} fine-tuning consistently increases the compression quality in terms of both automated metrics and the side-by-side evaluation (SbS score), especially for higher compression rates.
According to~\Cref{tab:time_embedding_ablation}, the effect of non-quantizing the timestep embedding layers is less significant.
Note that we omit the SbS scores for non-finetuned versions of $2$- and $3$-bit VQDM since their metrics and our preliminary analysis of the generated images shown that they were not competitive with the teacher model.

\vspace{-7pt}
\subsection{Distilled diffusion models}
\label{sec:sdxl_turbo}
\vspace{-3pt}
\begin{table*}[]
\footnotesize
\centering
\scriptsize
\setlength\tabcolsep{2.47pt}
\renewcommand{\arraystretch}{1.2}
\begin{tabular}{lcc|ccccc}
 \toprule
  \bf{Method} & \bf{Steps} & \bf{Avg bits} & \bf{Pickscore$\uparrow$} & \bf{CLIP$\uparrow$} & \bf{FID$\downarrow$} & \bf{SbS,\%$\uparrow$}\\
  \midrule
    \rowcolor[HTML]{dddddd} Original model & 4 & 32 & 0.229 & 0.355 & 24.49 & 50.0 \\
    VQDM & 4 & 4.15 & 0.228 & 0.355 & 24.89 & 43.9 \\
    VQDM & 4 & 3.15 & 0.226 & 0.352 & 26.55 &  29.9 \\
    \midrule
    \rowcolor[HTML]{dddddd} Original model & 2 & 32 & 0.230 & 0.357 & 26.43 & 50.0 \\
    VQDM & 2 & 4.15 & 0.230 & 0.357 & 26.43 & \cellcolor[HTML]{ccffcc} 46.5 \\
    VQDM & 2 & 3.15 & 0.230 & 0.357 & 26.59 & 30.1 \\
    \midrule
    \rowcolor[HTML]{dddddd} Original model & 1 & 32 & 0.228 & 0.359 & 26.07 & 50.0 \\
    VQDM & 1 & 4.15 & 0.226 & 0.357 & 26.30 & \cellcolor[HTML]{ccffcc} 45.5 \\
    VQDM & 1 & 3.15 & 0.221 & 0.346 & 30.64 & 26.6 \\
    \bottomrule
\end{tabular}
\caption{
    Evaluation of the quantized SDXL-Turbo model for different bit-widths and sampling steps. 
    Green indicates p-value $>0.1$ for the null hypothesis that the models perform equally.
}
\label{tab:sdxl_turbo}
\vspace{-12pt}
\end{table*}

\vspace{-5pt}
Finally, we apply our approach to SDXL-Turbo~\citep{sauer2023adversarial} operating in different sampling steps.
We present the results in~\Cref{tab:sdxl_turbo} and mark green the SbS scores providing p-value $>0.1$ for the null hypothesis that both models perform equally.
The results show that the distilled diffusion models can be compressed using VQDM to 4 bits without noticeable loss in performance.
\vspace{-5pt}

\vspace{-5pt}
\section{Conclusion}
\vspace{-3pt}

We introduce VQDM, a vector quantization method combined with a fine-tuning procedure aimed at compressing modern text-to-image diffusion models to low bit-widths. 
We take into account the specific architecture and inference process of diffusion models and adapt vector quantization to better suit these types of models. 
VQDM outperforms baseline methods in $3{-}4$ bit compression of SDXL.
Notably, our 3-bit compressed models perform on par with previous 4-bit quantization methods. 
Additionally, we demonstrate that vector quantization can be effectively applied to distilled diffusion models.

\bibliography{bibliography}

\begin{thebibliography}{84}
\providecommand{\natexlab}[1]{#1}
\providecommand{\url}[1]{\texttt{#1}}
\expandafter\ifx\csname urlstyle\endcsname\relax
  \providecommand{\doi}[1]{doi: #1}\else
  \providecommand{\doi}{doi: \begingroup \urlstyle{rm}\Url}\fi

\bibitem[AbouElhamayed et~al.(2023)AbouElhamayed, Cui, Fernandez-Marques, Lane, and Abdelfattah]{abouelhamayed2023pqa}
Ahmed~F AbouElhamayed, Angela Cui, Javier Fernandez-Marques, Nicholas~D Lane, and Mohamed~S Abdelfattah.
\newblock Pqa: Exploring the potential of product quantization in dnn hardware acceleration.
\newblock \emph{arXiv preprint arXiv:2305.18334}, 2023.

\bibitem[Babenko \& Lempitsky(2014)Babenko and Lempitsky]{aq}
Artem Babenko and Victor Lempitsky.
\newblock Additive quantization for extreme vector compression.
\newblock In \emph{Proceedings of the IEEE Conference on Computer Vision and Pattern Recognition}, pp.\  931--938, 2014.

\bibitem[Betker et~al.(2023)Betker, Goh, Jing, Brooks, Wang, Li, Ouyang, Zhuang, Lee, Guo, et~al.]{betker2023improving}
James Betker, Gabriel Goh, Li~Jing, Tim Brooks, Jianfeng Wang, Linjie Li, Long Ouyang, Juntang Zhuang, Joyce Lee, Yufei Guo, et~al.
\newblock Improving image generation with better captions.
\newblock \emph{Computer Science. https://cdn. openai. com/papers/dall-e-3. pdf}, 2\penalty0 (3):\penalty0 8, 2023.

\bibitem[{Black Forest Labs}(2024)]{flux}
{Black Forest Labs}.
\newblock Flux.1.
\newblock \url{https://huggingface.co/black-forest-labs/FLUX.1-dev}, 2024.

\bibitem[Blalock \& Guttag(2021)Blalock and Guttag]{blalock2021multiplying}
Davis Blalock and John Guttag.
\newblock Multiplying matrices without multiplying.
\newblock In \emph{International Conference on Machine Learning}, pp.\  992--1004. PMLR, 2021.

\bibitem[Burton et~al.(1983)Burton, Shore, and Buck]{vq1}
D.~Burton, J.~Shore, and J.~Buck.
\newblock A generalization of isolated word recognition using vector quantization.
\newblock In \emph{ICASSP '83. IEEE International Conference on Acoustics, Speech, and Signal Processing}, volume~8, pp.\  1021--1024, 1983.
\newblock \doi{10.1109/ICASSP.1983.1171915}.

\bibitem[Chang et~al.(2023)Chang, Shen, Cai, Ye, Xu, Cheng, Lv, Zhang, Lu, and Guo]{Chang2023EffectiveQF}
Hanwen Chang, Haihao Shen, Yiyang Cai, Xinyu Ye, Zhenzhong Xu, Wenhua Cheng, Kaokao Lv, Weiwei Zhang, Yintong Lu, and Heng Guo.
\newblock Effective quantization for diffusion models on cpus.
\newblock \emph{ArXiv}, abs/2311.16133, 2023.
\newblock URL \url{https://api.semanticscholar.org/CorpusID:265466086}.

\bibitem[Chee et~al.(2023)Chee, Cai, Kuleshov, and Sa]{chee2023quip}
Jerry Chee, Yaohui Cai, Volodymyr Kuleshov, and Christopher~De Sa.
\newblock Quip: 2-bit quantization of large language models with guarantees, 2023.

\bibitem[Chen et~al.(2023)Chen, Yu, Ge, Yao, Xie, Wu, Wang, Kwok, Luo, Lu, and Li]{chen2023pixartalpha}
Junsong Chen, Jincheng Yu, Chongjian Ge, Lewei Yao, Enze Xie, Yue Wu, Zhongdao Wang, James Kwok, Ping Luo, Huchuan Lu, and Zhenguo Li.
\newblock Pixart-$\alpha$: Fast training of diffusion transformer for photorealistic text-to-image synthesis, 2023.

\bibitem[Chen et~al.(2024)Chen, Wu, Luo, Xie, Paul, Luo, Zhao, and Li]{chen2024pixartdelta}
Junsong Chen, Yue Wu, Simian Luo, Enze Xie, Sayak Paul, Ping Luo, Hang Zhao, and Zhenguo Li.
\newblock Pixart-{$\delta$}: Fast and controllable image generation with latent consistency models, 2024.

\bibitem[Chen et~al.(2016)Chen, Xu, Zhang, and Guestrin]{chen2016training}
Tianqi Chen, Bing Xu, Chiyuan Zhang, and Carlos Guestrin.
\newblock Training deep nets with sublinear memory cost, 2016.

\bibitem[Chen et~al.(2010)Chen, Guan, and Wang]{rvq}
Yongjian Chen, Tao Guan, and Cheng Wang.
\newblock Approximate nearest neighbor search by residual vector quantization.
\newblock \emph{Sensors}, 10\penalty0 (12):\penalty0 11259--11273, 2010.

\bibitem[Dettmers et~al.(2023)Dettmers, Svirschevski, Egiazarian, Kuznedelev, Frantar, Ashkboos, Borzunov, Hoefler, and Alistarh]{dettmers2023spqr}
Tim Dettmers, Ruslan Svirschevski, Vage Egiazarian, Denis Kuznedelev, Elias Frantar, Saleh Ashkboos, Alexander Borzunov, Torsten Hoefler, and Dan Alistarh.
\newblock Spqr: A sparse-quantized representation for near-lossless llm weight compression.
\newblock \emph{arXiv preprint arXiv:2306.03078}, 2023.

\bibitem[Egiazarian et~al.(2024)Egiazarian, Panferov, Kuznedelev, Frantar, Babenko, and Alistarh]{egiazarian2024extreme}
Vage Egiazarian, Andrei Panferov, Denis Kuznedelev, Elias Frantar, Artem Babenko, and Dan Alistarh.
\newblock Extreme compression of large language models via additive quantization, 2024.

\bibitem[Esser et~al.(2024)Esser, Kulal, Blattmann, Entezari, Müller, Saini, Levi, Lorenz, Sauer, Boesel, Podell, Dockhorn, English, Lacey, Goodwin, Marek, and Rombach]{esser2024scaling}
Patrick Esser, Sumith Kulal, Andreas Blattmann, Rahim Entezari, Jonas Müller, Harry Saini, Yam Levi, Dominik Lorenz, Axel Sauer, Frederic Boesel, Dustin Podell, Tim Dockhorn, Zion English, Kyle Lacey, Alex Goodwin, Yannik Marek, and Robin Rombach.
\newblock Scaling rectified flow transformers for high-resolution image synthesis.
\newblock \emph{CoRR}, abs/2403.03206, 2024.

\bibitem[Esser et~al.(2020)Esser, McKinstry, Bablani, Appuswamy, and Modha]{Esser2020learned}
Steven~K. Esser, Jeffrey~L. McKinstry, Deepika Bablani, Rathinakumar Appuswamy, and Dharmendra~S. Modha.
\newblock Learned step size quantization.
\newblock In \emph{International Conference on Learning Representations}, 2020.
\newblock URL \url{https://openreview.net/forum?id=rkgO66VKDS}.

\bibitem[Fang et~al.(2023)Fang, Ma, and Wang]{fang2023structural}
Gongfan Fang, Xinyin Ma, and Xinchao Wang.
\newblock Structural pruning for diffusion models.
\newblock In \emph{Thirty-seventh Conference on Neural Information Processing Systems}, 2023.
\newblock URL \url{https://openreview.net/forum?id=d4f40zJJIS}.

\bibitem[Frantar et~al.(2022{\natexlab{a}})Frantar, Ashkboos, Hoefler, and Alistarh]{frantar-gptq}
Elias Frantar, Saleh Ashkboos, Torsten Hoefler, and Dan Alistarh.
\newblock {GPTQ}: Accurate post-training compression for generative pretrained transformers.
\newblock \emph{arXiv preprint arXiv:2210.17323}, 2022{\natexlab{a}}.

\bibitem[Frantar et~al.(2022{\natexlab{b}})Frantar, Singh, and Alistarh]{frantar2022obc}
Elias Frantar, Sidak~Pal Singh, and Dan Alistarh.
\newblock {Optimal Brain Compression:} a framework for accurate post-training quantization and pruning.
\newblock \emph{Advances in Neural Information Processing Systems}, 36, 2022{\natexlab{b}}.

\bibitem[Frantar et~al.(2023)Frantar, Ashkboos, Hoefler, and Alistarh]{Frantar2023OPTQ}
Elias Frantar, Saleh Ashkboos, Torsten Hoefler, and Dan Alistarh.
\newblock {OPTQ: Accurate Quantization for Generative Pre-trained Transformers}.
\newblock \emph{International Conference on Learning Representations (ICLR)}, 2023.

\bibitem[Gholami et~al.(2021)Gholami, Kim, Dong, Yao, Mahoney, and Keutzer]{gholami2021survey}
Amir Gholami, Sehoon Kim, Zhen Dong, Zhewei Yao, Michael~W. Mahoney, and Kurt Keutzer.
\newblock A survey of quantization methods for efficient neural network inference, 2021.

\bibitem[Goodfellow et~al.(2014)Goodfellow, Pouget-Abadie, Mirza, Xu, Warde-Farley, Ozair, Courville, and Bengio]{goodfellow2014generative}
Ian Goodfellow, Jean Pouget-Abadie, Mehdi Mirza, Bing Xu, David Warde-Farley, Sherjil Ozair, Aaron Courville, and Yoshua Bengio.
\newblock Generative adversarial nets.
\newblock In Z.~Ghahramani, M.~Welling, C.~Cortes, N.~Lawrence, and K.Q. Weinberger (eds.), \emph{Advances in Neural Information Processing Systems}, volume~27. Curran Associates, Inc., 2014.
\newblock URL \url{https://proceedings.neurips.cc/paper_files/paper/2014/file/5ca3e9b122f61f8f06494c97b1afccf3-Paper.pdf}.

\bibitem[Gray(1984)]{vq2}
R.~Gray.
\newblock Vector quantization.
\newblock \emph{IEEE ASSP Magazine}, 1\penalty0 (2):\penalty0 4--29, 1984.
\newblock \doi{10.1109/MASSP.1984.1162229}.

\bibitem[He et~al.(2016)He, Zhang, Ren, and Sun]{he2016residual}
Kaiming He, Xiangyu Zhang, Shaoqing Ren, and Jian Sun.
\newblock {Deep Residual Learning for Image Recognition}.
\newblock In \emph{Proceedings of 2016 IEEE Conference on Computer Vision and Pattern Recognition}, CVPR '16, pp.\  770--778. IEEE, June 2016.
\newblock \doi{10.1109/CVPR.2016.90}.
\newblock URL \url{http://ieeexplore.ieee.org/document/7780459}.

\bibitem[He et~al.(2023)He, Liu, Liu, Wu, Zhou, and Zhuang]{he2023ptqd}
Yefei He, Luping Liu, Jing Liu, Weijia Wu, Hong Zhou, and Bohan Zhuang.
\newblock {PTQD}: Accurate post-training quantization for diffusion models.
\newblock In \emph{Thirty-seventh Conference on Neural Information Processing Systems}, 2023.
\newblock URL \url{https://openreview.net/forum?id=Y3g1PV5R9l}.

\bibitem[He et~al.(2024)He, Liu, Wu, Zhou, and Zhuang]{he2024efficientdm}
Yefei He, Jing Liu, Weijia Wu, Hong Zhou, and Bohan Zhuang.
\newblock Efficient{DM}: Efficient quantization-aware fine-tuning of low-bit diffusion models.
\newblock In \emph{The Twelfth International Conference on Learning Representations}, 2024.
\newblock URL \url{https://openreview.net/forum?id=UmMa3UNDAz}.

\bibitem[Hessel et~al.(2021)Hessel, Holtzman, Forbes, Bras, and Choi]{clip}
Jack Hessel, Ari Holtzman, Maxwell Forbes, Ronan~Le Bras, and Yejin Choi.
\newblock Clipscore: {A} reference-free evaluation metric for image captioning.
\newblock \emph{CoRR}, abs/2104.08718, 2021.
\newblock URL \url{https://arxiv.org/abs/2104.08718}.

\bibitem[Heusel et~al.(2017)Heusel, Ramsauer, Unterthiner, Nessler, Klambauer, and Hochreiter]{fid}
Martin Heusel, Hubert Ramsauer, Thomas Unterthiner, Bernhard Nessler, G{\"{u}}nter Klambauer, and Sepp Hochreiter.
\newblock Gans trained by a two time-scale update rule converge to a nash equilibrium.
\newblock \emph{CoRR}, abs/1706.08500, 2017.
\newblock URL \url{http://arxiv.org/abs/1706.08500}.

\bibitem[Hinton et~al.(2015)Hinton, Vinyals, and Dean]{hinton2015distilling}
Geoffrey Hinton, Oriol Vinyals, and Jeff Dean.
\newblock Distilling the knowledge in a neural network.
\newblock \emph{arXiv preprint arXiv:1503.02531}, 2015.

\bibitem[Ho \& Salimans(2022)Ho and Salimans]{ho2022classifier}
Jonathan Ho and Tim Salimans.
\newblock Classifier-free diffusion guidance.
\newblock \emph{arXiv preprint arXiv:2207.12598}, 2022.

\bibitem[Ho et~al.(2020)Ho, Jain, and Abbeel]{ho2020denoising}
Jonathan Ho, Ajay Jain, and Pieter Abbeel.
\newblock Denoising diffusion probabilistic models.
\newblock \emph{Advances in neural information processing systems}, 33:\penalty0 6840--6851, 2020.

\bibitem[Huang et~al.(2024)Huang, Gong, Liu, Chen, and Liu]{huang2024tfmqdm}
Yushi Huang, Ruihao Gong, Jing Liu, Tianlong Chen, and Xianglong Liu.
\newblock Tfmq-dm: Temporal feature maintenance quantization for diffusion models.
\newblock In \emph{The IEEE/CVF Conference on Computer Vision and Pattern Recognition}, 2024.

\bibitem[Jegou et~al.(2010)Jegou, Douze, and Schmid]{pq}
Herve Jegou, Matthijs Douze, and Cordelia Schmid.
\newblock Product quantization for nearest neighbor search.
\newblock \emph{IEEE transactions on pattern analysis and machine intelligence}, 33\penalty0 (1):\penalty0 117--128, 2010.

\bibitem[Kang et~al.(2023)Kang, Zhu, Zhang, Park, Shechtman, Paris, and Park]{kang2023gigagan}
Minguk Kang, Jun-Yan Zhu, Richard Zhang, Jaesik Park, Eli Shechtman, Sylvain Paris, and Taesung Park.
\newblock Scaling up gans for text-to-image synthesis.
\newblock In \emph{Proceedings of the IEEE Conference on Computer Vision and Pattern Recognition (CVPR)}, 2023.

\bibitem[Kastryulin et~al.(2024)Kastryulin, Konev, Shishenya, Lyapustin, Khurshudov, Tselousov, Vinokurov, Kuznedelev, Markovich, Livshits, Kirillov, Tabisheva, Chubarova, Kaminskaia, Ustyuzhanin, Shvetsov, Shlenskii, Startsev, Kornilov, Romanov, Babenko, Ovcharenko, and Khrulkov]{kastryulin2024yaart}
Sergey Kastryulin, Artem Konev, Alexander Shishenya, Eugene Lyapustin, Artem Khurshudov, Alexander Tselousov, Nikita Vinokurov, Denis Kuznedelev, Alexander Markovich, Grigoriy Livshits, Alexey Kirillov, Anastasiia Tabisheva, Liubov Chubarova, Marina Kaminskaia, Alexander Ustyuzhanin, Artemii Shvetsov, Daniil Shlenskii, Valerii Startsev, Dmitrii Kornilov, Mikhail Romanov, Artem Babenko, Sergei Ovcharenko, and Valentin Khrulkov.
\newblock Yaart: Yet another art rendering technology, 2024.

\bibitem[Khan et~al.(2020)Khan, Sohail, Zahoora, and Qureshi]{khan2020survey}
Asifullah Khan, Anabia Sohail, Umme Zahoora, and Aqsa~Saeed Qureshi.
\newblock A survey of the recent architectures of deep convolutional neural networks.
\newblock \emph{Artificial intelligence review}, 53:\penalty0 5455--5516, 2020.

\bibitem[Kim et~al.(2023)Kim, Song, Castells, and Choi]{kim2023bksdm}
Bo-Kyeong Kim, Hyoung-Kyu Song, Thibault Castells, and Shinkook Choi.
\newblock Bk-sdm: Architecturally compressed stable diffusion for efficient text-to-image generation.
\newblock \emph{ICML Workshop on Efficient Systems for Foundation Models (ES-FoMo)}, 2023.
\newblock URL \url{https://openreview.net/forum?id=bOVydU0XKC}.

\bibitem[Kingma \& Welling(2022)Kingma and Welling]{kingma2022autoencoding}
Diederik~P Kingma and Max Welling.
\newblock Auto-encoding variational bayes, 2022.

\bibitem[Kirstain et~al.(2023)Kirstain, Polyak, Singer, Matiana, Penna, and Levy]{kirstain2023pickapic}
Yuval Kirstain, Adam Polyak, Uriel Singer, Shahbuland Matiana, Joe Penna, and Omer Levy.
\newblock Pick-a-pic: An open dataset of user preferences for text-to-image generation.
\newblock In \emph{Thirty-seventh Conference on Neural Information Processing Systems}, 2023.
\newblock URL \url{https://openreview.net/forum?id=G5RwHpBUv0}.

\bibitem[Li et~al.(2024)Li, Zou, Wang, Majumder, Xie, Manmatha, Swaminathan, Tu, Ermon, and Soatto]{li2024scalability}
Hao Li, Yang Zou, Ying Wang, Orchid Majumder, Yusheng Xie, R.~Manmatha, Ashwin Swaminathan, Zhuowen Tu, Stefano Ermon, and Stefano Soatto.
\newblock On the scalability of diffusion-based text-to-image generation, 2024.

\bibitem[Li et~al.(2023{\natexlab{a}})Li, Hu, Khan, Li, Yang, Wang, Cheng, and Yang]{li2023faster}
Senmao Li, Taihang Hu, Fahad~Shahbaz Khan, Linxuan Li, Shiqi Yang, Yaxing Wang, Ming-Ming Cheng, and Jian Yang.
\newblock Faster diffusion: Rethinking the role of unet encoder in diffusion models, 2023{\natexlab{a}}.

\bibitem[Li et~al.(2023{\natexlab{b}})Li, Liu, Lian, Yang, Dong, Kang, Zhang, and Keutzer]{li2023qdiffusion}
Xiuyu Li, Yijiang Liu, Long Lian, Huanrui Yang, Zhen Dong, Daniel Kang, Shanghang Zhang, and Kurt Keutzer.
\newblock Q-diffusion: Quantizing diffusion models.
\newblock In \emph{Proceedings of the IEEE/CVF International Conference on Computer Vision (ICCV)}, pp.\  17535--17545, October 2023{\natexlab{b}}.

\bibitem[Li et~al.(2023{\natexlab{c}})Li, Xu, Cao, Sun, and Zhang]{li2023qdm}
Yanjing Li, Sheng Xu, Xianbin Cao, Xiao Sun, and Baochang Zhang.
\newblock Q-{DM}: An efficient low-bit quantized diffusion model.
\newblock In \emph{Thirty-seventh Conference on Neural Information Processing Systems}, 2023{\natexlab{c}}.
\newblock URL \url{https://openreview.net/forum?id=sFGkL5BsPi}.

\bibitem[Li et~al.(2021)Li, Gong, Tan, Yang, Hu, Zhang, Yu, Wang, and Gu]{li2021brecq}
Yuhang Li, Ruihao Gong, Xu~Tan, Yang Yang, Peng Hu, Qi~Zhang, Fengwei Yu, Wei Wang, and Shi Gu.
\newblock {\{}BRECQ{\}}: Pushing the limit of post-training quantization by block reconstruction.
\newblock In \emph{International Conference on Learning Representations}, 2021.
\newblock URL \url{https://openreview.net/forum?id=POWv6hDd9XH}.

\bibitem[Lin et~al.(2024)Lin, Tang, Tang, Yang, Chen, Wang, Xiao, Dang, Gan, and Han]{lin2024awq}
Ji~Lin, Jiaming Tang, Haotian Tang, Shang Yang, Wei-Ming Chen, Wei-Chen Wang, Guangxuan Xiao, Xingyu Dang, Chuang Gan, and Song Han.
\newblock Awq: Activation-aware weight quantization for llm compression and acceleration, 2024.

\bibitem[Lin et~al.(2014)Lin, Maire, Belongie, Hays, Perona, Ramanan, Doll{\'a}r, and Zitnick]{lin2014microsoft}
Tsung-Yi Lin, Michael Maire, Serge Belongie, James Hays, Pietro Perona, Deva Ramanan, Piotr Doll{\'a}r, and C~Lawrence Zitnick.
\newblock Microsoft coco: Common objects in context.
\newblock In \emph{Computer Vision--ECCV 2014: 13th European Conference, Zurich, Switzerland, September 6-12, 2014, Proceedings, Part V 13}, pp.\  740--755. Springer, 2014.

\bibitem[Lu et~al.(2022)Lu, Zhou, Bao, Chen, Li, and Zhu]{lu2022dpm}
Cheng Lu, Yuhao Zhou, Fan Bao, Jianfei Chen, Chongxuan Li, and Jun Zhu.
\newblock Dpm-solver: A fast ode solver for diffusion probabilistic model sampling in around 10 steps.
\newblock \emph{arXiv preprint arXiv:2206.00927}, 2022.

\bibitem[Luo et~al.(2023)Luo, Tan, Huang, Li, and Zhao]{luo2023latent}
Simian Luo, Yiqin Tan, Longbo Huang, Jian Li, and Hang Zhao.
\newblock Latent consistency models: Synthesizing high-resolution images with few-step inference, 2023.

\bibitem[Ma et~al.(2024)Ma, Fang, and Wang]{ma2023deepcache}
Xinyin Ma, Gongfan Fang, and Xinchao Wang.
\newblock Deepcache: Accelerating diffusion models for free.
\newblock In \emph{The IEEE/CVF Conference on Computer Vision and Pattern Recognition}, 2024.

\bibitem[Martinez et~al.(2016)Martinez, Clement, Hoos, and Little]{lsq}
Julieta Martinez, Joris Clement, Holger~H Hoos, and James~J Little.
\newblock Revisiting additive quantization.
\newblock In \emph{Computer Vision--ECCV 2016: 14th European Conference, Amsterdam, The Netherlands, October 11-14, 2016, Proceedings, Part II 14}, pp.\  137--153. Springer, 2016.

\bibitem[Martinez et~al.(2018)Martinez, Zakhmi, Hoos, and Little]{lsq++}
Julieta Martinez, Shobhit Zakhmi, Holger~H Hoos, and James~J Little.
\newblock Lsq++: Lower running time and higher recall in multi-codebook quantization.
\newblock In \emph{Proceedings of the European Conference on Computer Vision (ECCV)}, pp.\  491--506, 2018.

\bibitem[Meng et~al.(2023)Meng, Rombach, Gao, Kingma, Ermon, Ho, and Salimans]{meng2023distillation}
Chenlin Meng, Robin Rombach, Ruiqi Gao, Diederik Kingma, Stefano Ermon, Jonathan Ho, and Tim Salimans.
\newblock On distillation of guided diffusion models.
\newblock In \emph{Proceedings of the IEEE/CVF Conference on Computer Vision and Pattern Recognition}, pp.\  14297--14306, 2023.

\bibitem[Micikevicius et~al.(2018)Micikevicius, Narang, Alben, Diamos, Elsen, Garcia, Ginsburg, Houston, Kuchaiev, Venkatesh, and Wu]{micikevicius2018mixed}
Paulius Micikevicius, Sharan Narang, Jonah Alben, Gregory Diamos, Erich Elsen, David Garcia, Boris Ginsburg, Michael Houston, Oleksii Kuchaiev, Ganesh Venkatesh, and Hao Wu.
\newblock Mixed precision training, 2018.

\bibitem[Nagel et~al.(2020)Nagel, Amjad, Van~Baalen, Louizos, and Blankevoort]{nagel20a}
Markus Nagel, Rana~Ali Amjad, Mart Van~Baalen, Christos Louizos, and Tijmen Blankevoort.
\newblock Up or down? {A}daptive rounding for post-training quantization.
\newblock In Hal~Daumé III and Aarti Singh (eds.), \emph{Proceedings of the 37th International Conference on Machine Learning}, volume 119 of \emph{Proceedings of Machine Learning Research}, pp.\  7197--7206. PMLR, 13--18 Jul 2020.

\bibitem[Nagel et~al.(2021{\natexlab{a}})Nagel, Fournarakis, Amjad, Bondarenko, van Baalen, and Blankevoort]{DBLP:journals/corr/abs-2106-08295}
Markus Nagel, Marios Fournarakis, Rana~Ali Amjad, Yelysei Bondarenko, Mart van Baalen, and Tijmen Blankevoort.
\newblock A white paper on neural network quantization.
\newblock \emph{CoRR}, abs/2106.08295, 2021{\natexlab{a}}.
\newblock URL \url{https://arxiv.org/abs/2106.08295}.

\bibitem[Nagel et~al.(2021{\natexlab{b}})Nagel, Fournarakis, Amjad, Bondarenko, van Baalen, and Blankevoort]{nagel2021white}
Markus Nagel, Marios Fournarakis, Rana~Ali Amjad, Yelysei Bondarenko, Mart van Baalen, and Tijmen Blankevoort.
\newblock A white paper on neural network quantization, 2021{\natexlab{b}}.

\bibitem[Nagel et~al.(2022)Nagel, Fournarakis, Bondarenko, and Blankevoort]{Nagel2022OvercomingOI}
Markus Nagel, Marios Fournarakis, Yelysei Bondarenko, and Tijmen Blankevoort.
\newblock Overcoming oscillations in quantization-aware training.
\newblock In \emph{International Conference on Machine Learning}, 2022.
\newblock URL \url{https://api.semanticscholar.org/CorpusID:247595112}.

\bibitem[Peebles \& Xie(2022)Peebles and Xie]{Peebles2022DiT}
William Peebles and Saining Xie.
\newblock Scalable diffusion models with transformers.
\newblock \emph{arXiv preprint arXiv:2212.09748}, 2022.

\bibitem[Pernias et~al.(2023)Pernias, Rampas, Richter, Pal, and Aubreville]{pernias2023wuerstchen}
Pablo Pernias, Dominic Rampas, Mats~L. Richter, Christopher~J. Pal, and Marc Aubreville.
\newblock Wuerstchen: An efficient architecture for large-scale text-to-image diffusion models, 2023.

\bibitem[Podell et~al.(2024)Podell, English, Lacey, Blattmann, Dockhorn, M{\"u}ller, Penna, and Rombach]{podell2024sdxl}
Dustin Podell, Zion English, Kyle Lacey, Andreas Blattmann, Tim Dockhorn, Jonas M{\"u}ller, Joe Penna, and Robin Rombach.
\newblock {SDXL}: Improving latent diffusion models for high-resolution image synthesis.
\newblock In \emph{The Twelfth International Conference on Learning Representations}, 2024.
\newblock URL \url{https://openreview.net/forum?id=di52zR8xgf}.

\bibitem[Rombach et~al.(2021)Rombach, Blattmann, Lorenz, Esser, and Ommer]{rombach2021highresolution}
Robin Rombach, Andreas Blattmann, Dominik Lorenz, Patrick Esser, and Björn Ommer.
\newblock High-resolution image synthesis with latent diffusion models, 2021.

\bibitem[Rombach et~al.(2022)Rombach, Blattmann, Lorenz, Esser, and Ommer]{rombach2022high}
Robin Rombach, Andreas Blattmann, Dominik Lorenz, Patrick Esser, and Bj{\"o}rn Ommer.
\newblock High-resolution image synthesis with latent diffusion models.
\newblock In \emph{Proceedings of the IEEE/CVF conference on computer vision and pattern recognition}, pp.\  10684--10695, 2022.

\bibitem[Ronneberger et~al.(2015)Ronneberger, Fischer, and Brox]{ronneberger2015u}
Olaf Ronneberger, Philipp Fischer, and Thomas Brox.
\newblock U-net: Convolutional networks for biomedical image segmentation.
\newblock In \emph{Medical image computing and computer-assisted intervention--MICCAI 2015: 18th international conference, Munich, Germany, October 5-9, 2015, proceedings, part III 18}, pp.\  234--241. Springer, 2015.

\bibitem[Saharia et~al.(2022)Saharia, Chan, Saxena, Li, Whang, Denton, Ghasemipour, Gontijo-Lopes, Ayan, Salimans, Ho, Fleet, and Norouzi]{saharia2022photorealistic}
Chitwan Saharia, William Chan, Saurabh Saxena, Lala Li, Jay Whang, Emily Denton, Seyed Kamyar~Seyed Ghasemipour, Raphael Gontijo-Lopes, Burcu~Karagol Ayan, Tim Salimans, Jonathan Ho, David~J. Fleet, and Mohammad Norouzi.
\newblock Photorealistic text-to-image diffusion models with deep language understanding.
\newblock In Alice~H. Oh, Alekh Agarwal, Danielle Belgrave, and Kyunghyun Cho (eds.), \emph{Advances in Neural Information Processing Systems}, 2022.
\newblock URL \url{https://openreview.net/forum?id=08Yk-n5l2Al}.

\bibitem[Sauer et~al.(2023{\natexlab{a}})Sauer, Karras, Laine, Geiger, and Aila]{Sauer2023ARXIV}
Axel Sauer, Tero Karras, Samuli Laine, Andreas Geiger, and Timo Aila.
\newblock {StyleGAN-T}: Unlocking the power of {GANs} for fast large-scale text-to-image synthesis.
\newblock volume abs/2301.09515, 2023{\natexlab{a}}.
\newblock URL \url{https://arxiv.org/abs/2301.09515}.

\bibitem[Sauer et~al.(2023{\natexlab{b}})Sauer, Lorenz, Blattmann, and Rombach]{sauer2023adversarial}
Axel Sauer, Dominik Lorenz, Andreas Blattmann, and Robin Rombach.
\newblock Adversarial diffusion distillation, 2023{\natexlab{b}}.

\bibitem[Sauer et~al.(2024)Sauer, Boesel, Dockhorn, Blattmann, Esser, and Rombach]{sauer2024fast}
Axel Sauer, Frederic Boesel, Tim Dockhorn, Andreas Blattmann, Patrick Esser, and Robin Rombach.
\newblock Fast high-resolution image synthesis with latent adversarial diffusion distillation, 2024.

\bibitem[Schuhmann et~al.(2022)Schuhmann, Beaumont, Vencu, Gordon, Wightman, Cherti, Coombes, Katta, Mullis, Wortsman, Schramowski, Kundurthy, Crowson, Schmidt, Kaczmarczyk, and Jitsev]{schuhmann2022laion5b}
Christoph Schuhmann, Romain Beaumont, Richard Vencu, Cade Gordon, Ross Wightman, Mehdi Cherti, Theo Coombes, Aarush Katta, Clayton Mullis, Mitchell Wortsman, Patrick Schramowski, Srivatsa Kundurthy, Katherine Crowson, Ludwig Schmidt, Robert Kaczmarczyk, and Jenia Jitsev.
\newblock Laion-5b: An open large-scale dataset for training next generation image-text models, 2022.

\bibitem[Shang et~al.(2023)Shang, Yuan, Xie, Wu, and Yan]{shang2023ptqdm}
Yuzhang Shang, Zhihang Yuan, Bin Xie, Bingzhe Wu, and Yan Yan.
\newblock Post-training quantization on diffusion models.
\newblock In \emph{CVPR}, 2023.

\bibitem[So et~al.(2023)So, Lee, Ahn, Kim, and Park]{so2023temporal}
Junhyuk So, Jungwon Lee, Daehyun Ahn, Hyungjun Kim, and Eunhyeok Park.
\newblock Temporal dynamic quantization for diffusion models.
\newblock In \emph{Thirty-seventh Conference on Neural Information Processing Systems}, 2023.
\newblock URL \url{https://openreview.net/forum?id=D1sECc9fiG}.

\bibitem[Sohl-Dickstein et~al.(2015)Sohl-Dickstein, Weiss, Maheswaranathan, and Ganguli]{sohl2015deep}
Jascha Sohl-Dickstein, Eric Weiss, Niru Maheswaranathan, and Surya Ganguli.
\newblock Deep unsupervised learning using nonequilibrium thermodynamics.
\newblock In \emph{International conference on machine learning}, pp.\  2256--2265. PMLR, 2015.

\bibitem[Song et~al.(2020)Song, Sohl-Dickstein, Kingma, Kumar, Ermon, and Poole]{song2020score}
Yang Song, Jascha Sohl-Dickstein, Diederik~P Kingma, Abhishek Kumar, Stefano Ermon, and Ben Poole.
\newblock Score-based generative modeling through stochastic differential equations.
\newblock \emph{arXiv preprint arXiv:2011.13456}, 2020.

\bibitem[Song et~al.(2023)Song, Dhariwal, Chen, and Sutskever]{song2023consistency}
Yang Song, Prafulla Dhariwal, Mark Chen, and Ilya Sutskever.
\newblock Consistency models.
\newblock \emph{arXiv preprint arXiv:2303.01469}, 2023.

\bibitem[Tang et~al.(2023)Tang, Wang, Chen, Guan, Wu, Tang, and Zhu]{tang2023posttraining}
Siao Tang, Xin Wang, Hong Chen, Chaoyu Guan, Zewen Wu, Yansong Tang, and Wenwu Zhu.
\newblock Post-training quantization with progressive calibration and activation relaxing for text-to-image diffusion models, 2023.

\bibitem[Tseng et~al.(2024)Tseng, Chee, Sun, Kuleshov, and Sa]{tseng2024quip}
Albert Tseng, Jerry Chee, Qingyao Sun, Volodymyr Kuleshov, and Christopher~De Sa.
\newblock Quip\#: Even better llm quantization with hadamard incoherence and lattice codebooks, 2024.

\bibitem[van Baalen et~al.(2024)van Baalen, Kuzmin, Nagel, Couperus, Bastoul, Mahurin, Blankevoort, and Whatmough]{vanbaalen2024gptvq}
Mart van Baalen, Andrey Kuzmin, Markus Nagel, Peter Couperus, Cedric Bastoul, Eric Mahurin, Tijmen Blankevoort, and Paul Whatmough.
\newblock Gptvq: The blessing of dimensionality for llm quantization, 2024.

\bibitem[Wang et~al.(2024)Wang, Shang, Yuan, Wu, and Yan]{wang2024quest}
Haoxuan Wang, Yuzhang Shang, Zhihang Yuan, Junyi Wu, and Yan Yan.
\newblock Quest: Low-bit diffusion model quantization via efficient selective finetuning, 2024.

\bibitem[Wang et~al.(2020)Wang, Chen, He, and Cheng]{10.5555/3524938.3525851}
Peisong Wang, Qiang Chen, Xiangyu He, and Jian Cheng.
\newblock Towards accurate post-training network quantization via bit-split and stitching.
\newblock In \emph{Proceedings of the 37th International Conference on Machine Learning}, ICML'20. JMLR.org, 2020.

\bibitem[Wimbauer et~al.(2024)Wimbauer, Wu, Schoenfeld, Dai, Hou, He, Sanakoyeu, Zhang, Tsai, Kohler, Rupprecht, Cremers, Vajda, and Wang]{wimbauer2024cache}
Felix Wimbauer, Bichen Wu, Edgar Schoenfeld, Xiaoliang Dai, Ji~Hou, Zijian He, Artsiom Sanakoyeu, Peizhao Zhang, Sam Tsai, Jonas Kohler, Christian Rupprecht, Daniel Cremers, Peter Vajda, and Jialiang Wang.
\newblock Cache me if you can: Accelerating diffusion models through block caching, 2024.

\bibitem[Yang et~al.(2023)Yang, Dai, Wang, Zhang, and Zhang]{yang2023efficient}
Yuewei Yang, Xiaoliang Dai, Jialiang Wang, Peizhao Zhang, and Hongbo Zhang.
\newblock Efficient quantization strategies for latent diffusion models, 2023.

\bibitem[Yu et~al.(2022)Yu, Xu, Koh, Luong, Baid, Wang, Vasudevan, Ku, Yang, Ayan, et~al.]{yu2022scaling}
Jiahui Yu, Yuanzhong Xu, Jing~Yu Koh, Thang Luong, Gunjan Baid, Zirui Wang, Vijay Vasudevan, Alexander Ku, Yinfei Yang, Burcu~Karagol Ayan, et~al.
\newblock Scaling autoregressive models for content-rich text-to-image generation.
\newblock \emph{arXiv preprint arXiv:2206.10789}, 2\penalty0 (3):\penalty0 5, 2022.

\bibitem[Zhao et~al.(2023)Zhao, Bai, Rao, Zhou, and Lu]{zhao2023unipc}
Wenliang Zhao, Lujia Bai, Yongming Rao, Jie Zhou, and Jiwen Lu.
\newblock Unipc: A unified predictor-corrector framework for fast sampling of diffusion models.
\newblock \emph{NeurIPS}, 2023.

\bibitem[Zhou et~al.(2024)Zhou, Chen, Wang, and Chen]{zhou2024fast}
Zhenyu Zhou, Defang Chen, Can Wang, and Chun Chen.
\newblock Fast ode-based sampling for diffusion models in around 5 steps, 2024.

\bibitem[Zhu et~al.(2024)Zhu, Zhang, Sifferman, Sheaves, Wang, Richmond, Zhou, and Eshraghian]{zhu2024scalable}
Rui-Jie Zhu, Yu~Zhang, Ethan Sifferman, Tyler Sheaves, Yiqiao Wang, Dustin Richmond, Peng Zhou, and Jason~K Eshraghian.
\newblock Scalable matmul-free language modeling.
\newblock \emph{arXiv preprint arXiv:2406.02528}, 2024.

\end{thebibliography}
\bibliographystyle{iclr2024_conference}

\appendix
\newpage
\appendix

\section{Implementation details} 
\label{app:implementation_details}
When applying this algorithm to large text-to-image models, we employ several tricks to make calibration more scalable. 
We use half-precision to quickly generate inference trajectories for the calibration set but run the rest of the calibration in full precision to avoid numeric issues.  
For fine-tuning, we opt for mixed precision ~\cite{micikevicius2018mixed} for forward and backward passes. 
On top of that, we use gradient checkpointing~\cite{chen2016training} and gradient accumulation to reduce the memory footprint during fine-tuning.

We use Adam optimizer for layer-wise calibration and fine-tuning with standard parameters $\beta_1$ and $\beta_2$. 
During fine-tuning, we found that using a sufficiently large batch size is crucial for the training to be effective. 
Namely, we fine-tune with at least 32 samples per batch and use 2048 generated samples for training.
The calibration dataset uses 256 prompts from the Pic-a-Pic dataset~\cite{kirstain2023pickapic}. 

For each quantization experiment, we used either the 2 NVIDIA H100 or 4 A100 GPUs. 
Quantizing the SDXL U-Net model took approximately 52h on 4 NVIDIA A100 and 36h on 2 NVIDIA H100. 
Fine-tuning the quantized model was conducted on 4 NVIDIA A100 and took from 8 to 28 hours, depending on the size of the fine-tuning dataset.

\section{Human evaluation setup}
\label{app:sbs}
\begin{figure*}[ht!]
    \centering
    \includegraphics[width=\linewidth]{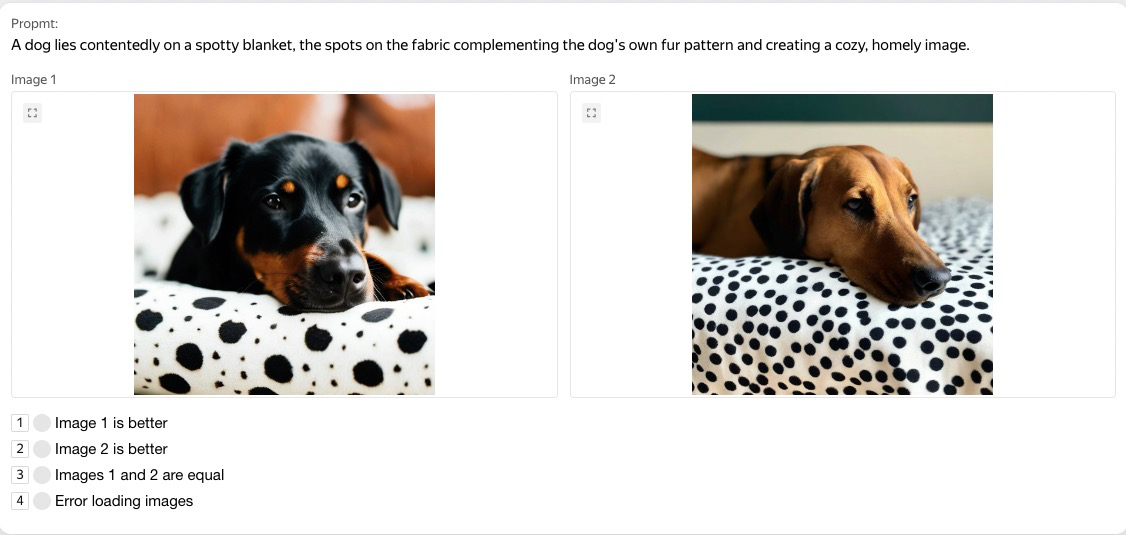}
    \caption{Side-by-side comparison interface for text-to-image human evaluation.}
    \label{fig:app_sbs}
\end{figure*}

The evaluation is conducted by professional assessors where they are asked to make a decision between two images given a textual prompt.
The decision is made according to image quality and textual alignment.
For each evaluated pair, three responses are collected and the final prediction is determined by majority voting.
We present the evaluation interface in Figure~\ref{fig:app_sbs}. 

\section{Limitations} 
\label{app:limitation}
Due to the limited computational resources and costly calibration and fine-tuning procedures for large diffusion models, we only experiment with two popular diffusion models: SDXL and SDXL Turbo that represent latent diffusion models based on the efficient U-Net architecture~\cite{podell2024sdxl}.
In general, VQDM is applicable to arbitrary diffusion architectures, including the transformer-based ones~\cite{Peebles2022DiT} that have also become widespread in practice~\cite{esser2024scaling, chen2023pixartalpha, chen2024pixartdelta}.
We leave the application of our method to the transformer-based~\cite{Peebles2022DiT} diffusion models for future work.
We also exclude the quantization of activations from the scope of our work, as our primary goal is to reduce memory requirements for modern diffusion models.

\section{Broader Impact} 
\label{app:broader_impact}
In this study, we propose a method for nearly lossless compression of diffusion models to $3-4$ bits, making the deployment of modern diffusion models feasible on general-purpose devices (e.g., gaming consoles or smartphones).
This paves the way for numerous new means to incorporate generative technology into everyday life, such as text-controlled photo editing or augmented reality applications based on diffusion models without sending data from the user device to the third-party server for computation.

On the other hand, as our method is applicable to all diffusion-based generative models, it also shares all the possible negative societal impacts of this technology, such as the generation of fake images and videos or the spread of social and cultural biases inherited by diffusion models.
However, we believe that our method does not introduce any new possibly harmful use cases for distributing text-to-image generation technology and would have a positive societal impact.

\section{Algorithms} 
\label{app:alg}

In this section, we describe both the quantization process (Algorithm~\ref{alg:quantization}) and the fine-tuning process (Algorithm~\ref{alg:finetuning}). 
First, the layers in the U-Net are quantized using Algorithm~\ref{alg:quantization}.
Then, we fine-tune the trainable weights of the quantized model to mimic the teacher's output (see Algorithm~\ref{alg:finetuning}).

\setlength{\textfloatsep}{0.3cm}
\begin{algorithm}[htb]
\caption{VQDM: Quantization stage.}
\label{alg:quantization}
\small
\begin{algorithmic}[1]
\vspace{-1px}\REQUIRE unet, prompt\_embed %
\FOR{$block \in \texttt{unet.blocks} $}
    \STATE $\mathbf{X}_{block} =collect\_block\_inputs(\texttt{unet},\texttt{prompt\_embed})$ \texttt{ // sampling with current unet}
    \FOR{$\texttt{layer}\ \ \texttt{in}\ \ \texttt{get\_linear\_and\_conv(block)}$}
        \STATE $\mathbf{W} := \texttt{layer.weight}$ 
        \STATE $\mathbf{X} := \texttt{get\_layer\_inputs}(\texttt{layer}, \mathbf{X}_{block})$
        \STATE $C, b, s := \texttt{initialize\_codebooks}(\mathbf{W})$ \texttt{ // k-means}
        \WHILE{ $\mathcal{L} > \tau$} %
            \STATE $C, s := \texttt{train\_Cs\_adam}(X, \mathbf{W}, C, b, s)$ %
            \STATE $b := \texttt{beam\_search}(X, \mathbf{W}, C, b, s)$ ~\citep{egiazarian2024extreme} 
        \ENDWHILE
        \STATE $\texttt{layer.weight} := \texttt{QuantizedLayer}(C, b, s)$
    \ENDFOR
\ENDFOR
\end{algorithmic}
\end{algorithm}

\setlength{\textfloatsep}{0.3cm}
\begin{algorithm}[h]
\caption{VQDM: Quantization-aware fine-tuning.}
\label{alg:finetuning}
\small
\begin{algorithmic}[1]
\vspace{-1px}\REQUIRE unet\_teacher, unet\_student, prompt\_embed
\STATE $\mathbf{dataset}:= get\_sampling\_trajectory(unet\_teacher, prompt\_embed)$

\FOR{$i = 1, \dots, \texttt{num\_epochs} $}
    \FOR{$latents\_batch\ \ \texttt{in}\ \ \texttt{dataset} $}
        
        \STATE $\texttt{calculate\_loss}(unet\_teacher, unet\_student,latents\_batch, prompt\_embed)$
        \STATE $optimize(unet\_student)$
    \ENDFOR
\ENDFOR
\end{algorithmic}
    \texttt{func} $\textbf{calculate\_loss}$
\begin{algorithmic}[1]
\vspace{-1px}\REQUIRE unet\_teacher, unet\_student, latents, prompt\_embeds
    \STATE $\mathbf{Y}_{teacher} =\texttt{unet\_teacher}(latents, prompt\_embeds)$
    \STATE $\mathbf{Y}_{student} =\texttt{unet\_student}(latents\_batch, prompt\_embeds)$
    \STATE $loss = ||\mathbf{Y}_{student} - \mathbf{Y}_{teacher}||^2_2$

\end{algorithmic}
\end{algorithm}

\end{document}